\documentclass[conference]{IEEEtran}

\IEEEoverridecommandlockouts

\usepackage{cite}
\usepackage{amsmath,amssymb,amsfonts}
\usepackage{algorithmic}
\usepackage{graphicx}
\usepackage{textcomp}
\usepackage{xcolor}
\usepackage{tikz}
\usetikzlibrary{positioning,arrows.meta,fit,calc,shapes.geometric}
\usepackage{newtxtext}
\usepackage[hidelinks]{hyperref}
\makeatletter
\renewcommand\section{\@startsection{section}{1}{\z@}%
  {1.5ex plus 1ex minus .2ex}%
  {1ex plus .2ex}%
  {\centering\normalfont\bfseries}}
\makeatother

\begin{document}

\title{
    Muon Meets Mamba: \\
    Spectral Optimization for State Space Models
}

\author{
    \IEEEauthorblockN{Arslan Battalov}
    \IEEEauthorblockA{
        \textit{HSE University} \\
        Moscow, Russia
    }
    \and
    \IEEEauthorblockN{Karim Kramin}
    \IEEEauthorblockA{
        \textit{HSE University} \\
        Moscow, Russia
    }
    \and
    \IEEEauthorblockN{Alexander Markotenko}
    \IEEEauthorblockA{
        \textit{HSE University} \\
        Moscow, Russia
    }
    \and
    \IEEEauthorblockN{Sofia Sinitsina}
    \IEEEauthorblockA{
        \textit{HSE University} \\
        Moscow, Russia
    }
    \thanks{All authors contributed equally. Author names are listed
    alphabetically.}
}

\maketitle

\begin{abstract}
Muon is a recent optimizer that orthogonalizes the update to each weight
matrix with a Newton--Schulz iteration, which performs steepest descent
under the spectral norm. Almost all the evidence for it comes from
Transformer models, and its behavior on state-space models is largely
unreported. We compare Muon with AdamW on Mamba-2 130M under a controlled
protocol that varies only which weight groups are trained with Muon. The
benefit is localized. Muon on the output projection alone beats Muon on
the input projection or on both. The advantage is mainly one of token
efficiency. It holds on two corpora and two token budgets, and persists
when training continues well past the compute-optimal point. Conditioning
does not explain the gain. Muon lowers the condition number of whichever
projection it trains, but the better-conditioned input projection is not
the one that helps.
\end{abstract}


\section{Project Description}

\begin{figure*}[t]
\centering
\begin{tikzpicture}[
  node distance=2.2mm and 6mm,
  block/.style={rectangle, draw, minimum height=5mm, inner sep=2.5pt,
                font=\scriptsize, align=center, rounded corners=1pt},
  noeleg/.style={block, fill=black!7, draw=black!45},
  gin/.style={block, fill=blue!18, line width=0.6pt, draw=blue!55!black},
  gout/.style={block, fill=green!22, line width=0.6pt, draw=green!45!black},
  txt/.style={font=\scriptsize, align=center},
  arr/.style={-{Stealth[length=1.3mm]}, line width=0.4pt}
]

\node[txt] (in) {input $u_t \in \mathbb{R}^{d_{\mathrm{model}}}$};

\node[gin, below=of in, minimum width=6.0cm] (inproj)
  {\texttt{in\_proj}$_\ell$\texttt{.weight}
   \,\,$\in \mathcal{G}_{\mathrm{in}}$
   \,\,(Muon at $k\!\in\!\{1,2\}$)};

\node[txt, text width=6.0cm, below=1mm of inproj] (split)
  {row split: $z$, $x,B,C$, $\Delta t$};

\node[noeleg, below=1mm of split, minimum width=6.0cm] (conv)
  {\texttt{conv1d}$_\ell$\texttt{.weight} (depthwise, 3-D tensor)};

\node[noeleg, below=of conv, minimum width=6.0cm] (ssd)
  {SSD core: \texttt{A\_log}$_\ell$, \texttt{D}$_\ell$,
   \texttt{dt\_bias}$_\ell$ (1-D)};

\node[noeleg, below=of ssd, minimum width=6.0cm] (gate)
  {gating $y \leftarrow z \odot y_{\mathrm{ssd}}$};

\node[noeleg, below=of gate, minimum width=6.0cm] (norm)
  {\texttt{norm}$_\ell$\texttt{.weight} (1-D RMSNorm)};

\node[gout, below=of norm, minimum width=6.0cm] (outproj)
  {\texttt{out\_proj}$_\ell$\texttt{.weight}
   \,\,$\in \mathcal{G}_{\mathrm{out}}$
   \,\,(Muon at $k\!\in\!\{2,3\}$)};

\node[txt, below=of outproj] (out)
  {output $y_t \in \mathbb{R}^{d_{\mathrm{model}}}$};

\draw[arr] (in)     -- (inproj);
\draw[arr] (inproj) -- (split);
\draw[arr] (split)  -- (conv);
\draw[arr] (conv)   -- (ssd);
\draw[arr] (ssd)    -- (gate);
\draw[arr] (gate)   -- (norm);
\draw[arr] (norm)   -- (outproj);
\draw[arr] (outproj)-- (out);

\end{tikzpicture}
\caption{Mamba-2 block (single layer~$\ell$) with the
two-dimensional, Muon-eligible projection matrices highlighted.
Muon is assigned to blue $\mathcal{G}_{\mathrm{in}}$ at $k=1$ and
$k=2$, and to green $\mathcal{G}_{\mathrm{out}}$ at $k=2$ (jointly with
$\mathcal{G}_{\mathrm{in}}$) and at $k=3$ (alone). Gray cells are parameters always trained
with AdamW: one-dimensional SSD and normalization parameters,
three-dimensional depthwise convolution filters, embeddings, the
language-modeling head, and biases.}
\label{fig:block}
\end{figure*}
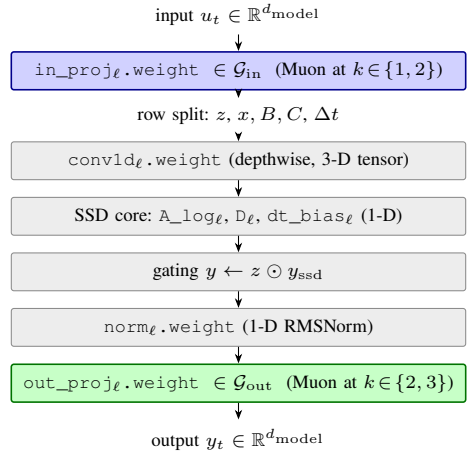

We compare how Mamba-2 trains under Muon versus AdamW in a controlled
setup. The model is Mamba-2 130M, trained on two datasets, OpenWebText
(OWT) and FineWeb-Edu, at token budgets of \(10^9\), \(2.6\times10^9\),
and \(5\times10^{10}\). The baseline trains every parameter with AdamW.
Separate runs, from the same initialization seeds, instead assign Muon
to \(\mathcal{G}_{\mathrm{in}}\), to \(\mathcal{G}_{\mathrm{out}}\), and
to \(\mathcal{G}_{\mathrm{in}}\cup\mathcal{G}_{\mathrm{out}}\), and leave
the rest on AdamW. Section~\ref{sec:background} describes the model and
the two optimizers, Section~\ref{sec:problem} defines these assignments
formally, and Figure~\ref{fig:block} shows them inside the Mamba-2 block.

The central research question is whether Muon, applied to
\(\mathcal{G}_{\mathrm{in}}\), to \(\mathcal{G}_{\mathrm{out}}\), or to
\(\mathcal{G}_{\mathrm{in}}\cup\mathcal{G}_{\mathrm{out}}\), improves on
the pure AdamW baseline on Mamba-2, in final validation loss or in token
efficiency, and whether any such improvement is localized to a specific
projection group.

All runs use the official Mamba implementation
\cite{statespaces2024mamba} and the public Muon reference implementation
\cite{jordan2024muonblog,kellerjordan2024muonrepo}. The contribution is
empirical: we report final loss, convergence, and robustness under one
fixed protocol, together with the matrix spectral diagnostics used to
interpret where the Muon assignment helps.

\section{Literature Review}
\label{sec:literature}

\paragraph*{Mamba-2 as the target architecture}
Structured state space models such as S4 \cite{gu2022efficiently}
showed that recurrent sequence models with structured transition
matrices can be competitive on long-range sequence tasks. Mamba
\cite{gu2023mamba} made this family effective for language modeling by
introducing input-dependent selective SSM parameters. Mamba-2
\cite{dao2024mamba2} reformulates the sequence-mixing computation through
structured state space duality, which connects SSMs to structured matrices
and linear attention, and improves hardware utilization relative to
Mamba. Empirical studies show that Mamba-family models behave
differently from Transformers in architecture-specific ways: Waleffe et al.\
\cite{waleffe2024empirical} compare pure SSM and hybrid models up to 8B
parameters, Park et al.\ \cite{park2024can} analyze controlled
in-context learning tasks, and Lieber et al.\ \cite{lieber2024jamba}
train Jamba, a 52B-parameter hybrid Transformer--Mamba model. This
architecture-specific behavior is why optimizer conclusions from
Transformers cannot be assumed to transfer, and why we treat Mamba-2
as a separate optimization target.

\paragraph*{AdamW as the baseline optimizer}
AdamW \cite{loshchilov2019adamw} decouples weight decay from the adaptive
gradient update and is the baseline optimizer here; all parameters not
assigned to Muon remain on AdamW. Recent work explains why Adam-type
methods are strong baselines for language modeling. Zhang et al.\
\cite{zhang2024transformers} identify blockwise Hessian heterogeneity as
one reason a single global learning rate can make SGD underperform Adam
on Transformers. Kunstner et al.\ \cite{kunstner2024heavy} show that
heavy-tailed token imbalance slows gradient descent on low-frequency
classes, while Adam and sign-based methods are less sensitive. Zhao et
al.\ \cite{zhao2024deconstructing} provide the optimizer-comparison
template used here: fixed token budgets, explicit learning-rate
selection, and a strong baseline. We therefore compare Muon
against a carefully calibrated AdamW baseline under a fixed protocol.

\paragraph*{Matrix-aware optimization and Muon}
Shampoo \cite{anil2020scalable,shi2023distributed} is the main precedent
here. It preconditions whole weight matrices with Kronecker factors
rather than adapting each coordinate independently, but requires inverse
matrix-root computations and substantial optimizer-state engineering.
Muon is lighter, as it applies a fixed-iteration
Newton--Schulz orthogonalization directly to matrix-valued momentum
updates. Jordan \cite{jordan2024muonblog} defines Muon for
two-dimensional hidden-layer matrices, while embeddings, output matrices,
scalars, and vectors remain under AdamW. Liu et al.\ \cite{liu2025muon}
report that, with weight decay and shape-dependent update scaling, Muon
matches AdamW quality at approximately \(52\%\) of the training FLOPs in
Transformer-based scaling-law experiments. Bernstein and Newhouse
\cite{bernstein2024old,bernstein2024modular} connect orthogonalized
matrix updates to steepest descent under the spectral norm and argue that
the appropriate update geometry depends on layer role. The role of
Mamba-2's projection matrices differs from that of Transformer attention
or MLP projections, so Muon's behavior on them is not predetermined by
the Transformer evidence.

\paragraph*{Adjacent formulations}
Two kinds of alternatives sit next to Muon. Per-group AdamW learning
rates and plain momentum methods probe learning-rate allocation and
momentum, but not Muon's spectral update geometry. Shampoo and related
preconditioners do reach that geometry through explicit preconditioning,
but at the cost of larger optimizer state and more implementation
work~\cite{anil2020scalable,shi2023distributed}. We therefore use Muon,
the lightweight option, and vary only its assignment to the predefined
Mamba-2 projection groups.

\paragraph*{Spectral diagnostics}
Muon orthogonalizes matrix-valued update directions, so its effect may
appear in the spectra of the trained matrices to which it is assigned.
Saxe, McClelland, and Ganguli \cite{saxe2013exact} show in deep linear
networks that learning dynamics decompose along singular modes and that
ill-conditioned matrices can slow learning by separating the evolution
of different modes. Martin and Mahoney \cite{martin2021implicit} find
that trained deep-network weight matrices can exhibit structured
spectral regimes, including spiked and heavy-tailed spectra, rather than
uniform singular-value distributions. Together these results make the
spectral effect of Muon on Mamba-2 projection matrices hard to predict.
A more uniform update geometry may help if AdamW leaves the projection
matrices poorly conditioned, but it could equally be neutral or harmful
if the AdamW-trained spectra already carry useful task-specific structure.
We therefore report three diagnostics (effective rank
\cite{roy2007effective}, condition number, and spectral norm) to test
whether optimizer-dependent validation differences come with systematic
spectral changes in the Mamba-2 projection groups.

\paragraph*{Research gap}
No controlled evidence exists for how Muon should be assigned inside
Mamba-2 blocks. Existing work supports Muon as a spectral-geometry update
for Transformer hidden matrices, but says nothing about which Mamba-2
projection matrices should receive it. We close that gap by comparing
AdamW and Muon under a shared protocol, varying only which predefined
Mamba-2 projection groups receive Muon.
\section{Preliminaries}
\label{sec:background}

This section fixes the model and the two optimizers, and identifies
which Mamba-2 parameters are matrix-valued and therefore eligible for
Muon.

\paragraph*{Model and SSD core}
Let \(\Theta\) be the trainable parameters of the Mamba-2 130M model
(Section~\ref{sec:metrics}) and \(L\) the number of Mamba-2 blocks.
Following Dao and Gu \cite{dao2024mamba2}, a discrete SSM has the form
\[
    h_t=A_t h_{t-1}+B_t x_t,
    \qquad
    y_t=C_t^\top h_t.
\]
In the default Mamba-2 multi-input-head pattern, and suppressing the
batch axis, the SSD layer operates on
\[
    A \in \mathbb{R}^{T \times H},
    \qquad
    X \in \mathbb{R}^{T \times H \times P},
    \qquad
    B,C \in \mathbb{R}^{T \times 1 \times N},
\]
where \(T\) is the sequence length, \(H\) the number of heads,
\(P\) the head dimension, and \(N\) the state dimension.

The SSD layer maps \(A,X,B,C\) to \(Y\), but those quantities come
from trainable block projections and nonlinearities; they are not
themselves the matrices Muon acts on.

\paragraph*{Eligible matrices in a Mamba-2 block}
For the reference state-spaces/mamba implementation
\cite{statespaces2024mamba} of a single Mamba-2 block, indexed by
layer \(\ell\in\{1,\dots,L\}\), the trainable parameters comprise:
\begin{itemize}
    \item \texttt{in\_proj}\(_\ell\)\texttt{.weight}
    \(\in\mathbb{R}^{(2 d_{\mathrm{inner}}
    + 2 n_{\mathrm{groups}} d_{\mathrm{state}} + n_{\mathrm{heads}})
    \times d_{\mathrm{model}}}\),
    the merged input projection that produces the gating channel
    \(z\), the data channel \(x\), the SSD observations
    \(B,C\), and the time-step \(\Delta t\);
    \item \texttt{conv1d}\(_\ell\)\texttt{.weight} and its bias
    \texttt{conv1d}\(_\ell\)\texttt{.bias}, a depthwise 1D
    convolution over the \((x,B,C)\) channels (a 3-dimensional
    tensor, not a matrix in the linear-algebra sense);
    \item \texttt{dt\_bias}\(_\ell\), \texttt{A\_log}\(_\ell\),
    \texttt{D}\(_\ell\) \(\in\mathbb{R}^{n_{\mathrm{heads}}}\),
    one-dimensional SSD parameters;
    \item the block input norm \texttt{norm}\(_\ell\)\texttt{.weight}
    \(\in\mathbb{R}^{d_{\mathrm{model}}}\) and the gated norm
    \texttt{mixer.norm}\(_\ell\)\texttt{.weight}
    \(\in\mathbb{R}^{d_{\mathrm{inner}}}\), both RMSNorm weights;
    \item \texttt{out\_proj}\(_\ell\)\texttt{.weight}
    \(\in\mathbb{R}^{d_{\mathrm{model}}\times d_{\mathrm{inner}}}\),
    the output projection.
\end{itemize}
The two-dimensional matrix-valued trainable parameters in this
block are therefore \texttt{in\_proj}\(_\ell\)\texttt{.weight} and
\texttt{out\_proj}\(_\ell\)\texttt{.weight}. Token embeddings, the
language-modeling head, biases, normalization parameters,
one-dimensional SSD parameters, and the depthwise convolution
filter are excluded from Muon assignment by the eligibility rules
of \cite{jordan2024muonblog,liu2025muon} and remain on AdamW
throughout this study.

We accordingly define
\[
    \mathcal{G}_{\mathrm{in}}
    =
    \{\texttt{in\_proj}_\ell\texttt{.weight} : \ell = 1,\dots,L\},
\]
\[
    \mathcal{G}_{\mathrm{out}}
    =
    \{\texttt{out\_proj}_\ell\texttt{.weight} : \ell = 1,\dots,L\}.
\]
For the Mamba-2 architecture used in the main study, the
Muon-eligible matrix set inside the blocks is
\[
    \mathcal{G}_{\mathrm{elig}}
    =
    \mathcal{G}_{\mathrm{in}}
    \cup
    \mathcal{G}_{\mathrm{out}} .
\]
All parameters in
\(\Theta\setminus\mathcal{G}_{\mathrm{elig}}\) are always trained
with AdamW.

\paragraph*{Optimizer update rules}
For the minibatch training loss \(\mathcal{L}(\theta_n)\) at update
step \(n\), let
\[
    g_n
    =
    \nabla_\theta \mathcal{L}(\theta_n)
\]
denote the full parameter gradient. We write \(g_{n,w}\) and
\(g_{n,W}\) for the components of \(g_n\) at a parameter \(w\) and at a
matrix-valued parameter \(W\), respectively.

AdamW uses the decoupled weight decay of Loshchilov and Hutter
\cite{loshchilov2019adamw}. For a parameter \(w\) on AdamW, it
maintains elementwise moment estimates
\[
    m_{n,w}
    =
    \beta_1 m_{n-1,w}
    +
    (1-\beta_1) g_{n,w},
\]
\[
    v_{n,w}
    =
    \beta_2 v_{n-1,w}
    +
    (1-\beta_2)(g_{n,w} \odot g_{n,w}),
\]
with bias-corrected moments
\[
    \widehat m_{n,w}
    =
    \frac{m_{n,w}}{1-\beta_1^n},
    \qquad
    \widehat v_{n,w}
    =
    \frac{v_{n,w}}{1-\beta_2^n}.
\]
The AdamW update is
\[
    w_{n+1}
    =
    w_n
    -
    \alpha
    \frac{\widehat m_{n,w}}{\sqrt{\widehat v_{n,w}}+\varepsilon}
    -
    \alpha \lambda w_n .
\]
We use betas \((\beta_1,\beta_2)=(0.9,0.95)\),
\(\varepsilon=10^{-8}\), and decoupled weight decay
\(\lambda=0.1\). Following the configuration used by Dao and Gu
\cite{dao2024mamba2} for Mamba-2 130M, the AdamW \emph{reference}
learning rate is
\[
    \alpha_0 = 3\times 10^{-3}.
\]

On Muon-updated parameters, for each matrix-valued parameter
\(W\in\mathcal{G}_{\mathrm{elig}}\), the optimizer forms a matrix-valued
momentum buffer
\[
    M_{n,W} = \mu\, M_{n-1,W} + (1-\mu)\, g_{n,W},
    \qquad \mu = 0.95,
\]
an exponential moving average of the gradient. Following the public
Muon reference implementation
\cite{jordan2024muonblog,kellerjordan2024muonrepo}, the buffer is
combined with the current gradient in a Nesterov step,
\[
    U_{n,W} = (1-\mu)\, g_{n,W} + \mu\, M_{n,W},
\]
and it is \(U_{n,W}\), rather than \(M_{n,W}\) itself, that is
approximately orthogonalized by the Newton--Schulz polynomial of
\cite{jordan2024muonblog,liu2025muon}.
Letting
\[
    \widetilde U_{n,W}
    =
    \frac{U_{n,W}}{\lVert U_{n,W}\rVert_F}
\]
be the Frobenius-normalized update direction, the iteration is
initialized at \(X_0 = \widetilde U_{n,W}\) and applies, for
\(j = 0,1,\dots,N_{\mathrm{NS}}-1\),
\[
    X_{j+1}
    =
    a\, X_j
    +
    b\, X_j X_j^{\!\top} X_j
    +
    c\, X_j X_j^{\!\top} X_j X_j^{\!\top} X_j,
\]
with the quintic coefficients
\[
    (a,b,c)=(3.4445,\,-4.7750,\,2.0315),
\]
reported by \cite{jordan2024muonblog} and used as the default in
the public Muon reference implementation
\cite{kellerjordan2024muonrepo}. The orthogonalized matrix is then
rescaled by
\(\gamma_W = \max(1,\, d^{\mathrm{row}}_W / d^{\mathrm{col}}_W)^{1/2}\),
where \(d^{\mathrm{row}}_W \times d^{\mathrm{col}}_W\) is the shape of
\(W\), a shape-dependent factor that keeps the update magnitude
comparable to AdamW's across matrices of different aspect ratios. This
default scaling is distinct from the RMS-matched rescaling evaluated as
an ablation in Appendix~\ref{sec:rms-wd-ablations}, which is not used in
the main protocol.
With learning rate \(\eta\), the Muon update is
\[
    W_{n+1}
    =
    W_n
    -
    \eta\, \gamma_W\, X_{N_{\mathrm{NS}}}.
\]
The polynomial \(p(\sigma)=a\sigma+b\sigma^3+c\sigma^5\) pushes the
singular values of \(\widetilde U_{n,W}\) toward \(1\), so
\(X_{N_{\mathrm{NS}}}\) approximates the orthogonal polar factor of
\(U_{n,W}\) and has approximately unit spectral norm; the resulting
step is spectral rather than elementwise, and is therefore not
commensurate with AdamW's adaptive update. We use
\(N_{\mathrm{NS}}=5\) iterations, following
\cite{jordan2024muonblog,liu2025muon}. The Muon
\emph{reference} learning rate is
\[
    \eta_0 = 1\times 10^{-2},
\]
within the range \([5\times 10^{-3},\,2\times 10^{-2}]\) used in
\cite{jordan2024muonblog,liu2025muon}.

The reference rates \(\alpha_0\) and \(\eta_0\) only fix the search
neighborhood: the learning rates actually used in all runs are
selected by the pilot calibration of Section~\ref{sec:lr-calibration}.

\section{Problem Statement}
\label{sec:problem}

\paragraph*{Autoregressive language modeling}
We consider autoregressive language modeling over a finite vocabulary
\(\mathcal{V}\). For a token sequence
\(x=(x_1,\dots,x_T)\in\mathcal{V}^T\) and model parameters \(\theta\),
the model defines the conditional distribution
\[
    p_\theta(x)=\prod_{t=1}^{T}p_\theta(x_t \mid x_{<t}),
    \qquad x_{<1}=\varnothing .
\]
Training minimizes the average negative log-likelihood on
\(\mathcal{D}_{\mathrm{train}}\), and model quality is evaluated on a
held-out validation set \(\mathcal{D}_{\mathrm{val}}\) by
\[
    \mathcal{L}_{\mathrm{val}}(\theta)
    =
    -\frac{1}{N_{\mathrm{val}}}
    \sum_{x\in\mathcal{D}_{\mathrm{val}}}
    \sum_{t=1}^{|x|}
    \log p_\theta(x_t\mid x_{<t}),
\]
\[
    N_{\mathrm{val}}
    =
    \sum_{x\in\mathcal{D}_{\mathrm{val}}}|x|.
\]
The corresponding validation perplexity is
\[
    \mathrm{PPL}_{\mathrm{val}}(\theta)
    =
    \exp\!\bigl(\mathcal{L}_{\mathrm{val}}(\theta)\bigr).
\]

\paragraph*{Muon assignment regimes}
The Muon-eligible matrix groups \(\mathcal{G}_{\mathrm{in}}\),
\(\mathcal{G}_{\mathrm{out}}\), and the eligible set
\(\mathcal{G}_{\mathrm{elig}}=\mathcal{G}_{\mathrm{in}}\cup\mathcal{G}_{\mathrm{out}}\)
are defined in Section~\ref{sec:background}. With two eligible groups,
Muon can be assigned to neither, to one of them, or to both, which
enumerates four regimes
\[
    \begin{aligned}
        \mathcal{M}^{(0)} &= \varnothing, \\
        \mathcal{M}^{(1)} &= \mathcal{G}_{\mathrm{in}}, \\
        \mathcal{M}^{(2)} &=
        \mathcal{G}_{\mathrm{in}}
        \cup
        \mathcal{G}_{\mathrm{out}}, \\
        \mathcal{M}^{(3)} &= \mathcal{G}_{\mathrm{out}} .
    \end{aligned}
\]
For each \(k\in\{0,1,2,3\}\), the parameters in \(\mathcal{M}^{(k)}\)
are trained with Muon and the rest, \(\Theta\setminus\mathcal{M}^{(k)}\),
with AdamW. Thus \(k=0\) is the pure AdamW baseline, \(k=1\) puts Muon on
the merged input projections, \(k=3\) on the output projections, and
\(k=2\) on both. Here \(\mathcal{M}^{(2)}=\mathcal{G}_{\mathrm{elig}}\)
means Muon on all eligible matrix-valued parameters inside the Mamba-2
blocks, not on every trainable parameter of the model. Each regime is
trained from scratch from the same shared seed initialization; no run is
warm-started from an AdamW checkpoint.

We select each optimizer's learning rate separately and then hold it
fixed across regimes; the controlled experimental factor is the Muon
assignment to the fixed sets \(\mathcal{M}^{(k)}\).

\paragraph*{Research question}
We ask three questions, with the assignment regime \(k\) as the only
controlled factor. We answer them mainly with the validation loss
\(\mathcal{L}_{\mathrm{val}}\) and perplexity
\(\mathrm{PPL}_{\mathrm{val}}\) defined above, and also report token
efficiency and spectral diagnostics:
\begin{enumerate}
    \item does assigning Muon to the input projections (\(k=1\)), the
    output projections (\(k=3\)), or both (\(k=2\)) reduce the final
    validation loss relative to the pure AdamW baseline (\(k=0\)) on
    Mamba-2;
    \item is any improvement \emph{localized} to a particular projection
    group, rather than shared equally between
    \(\mathcal{G}_{\mathrm{in}}\) and \(\mathcal{G}_{\mathrm{out}}\);
    \item is the observed effect stable across the training corpus and
    the token budget.
\end{enumerate}

\section{Experimental Protocol and Evaluation Metrics}
\label{sec:metrics}

We train Mamba-2 130M under the regimes of Section~\ref{sec:problem} on
two datasets, OpenWebText (OWT)~\cite{gokaslan2019openwebtext} and
FineWeb-Edu (FW)~\cite{penedo2024fineweb}, at token budgets \(10^9\) and
the Chinchilla-optimal \(2.6\times10^9\)~\cite{hoffmann2022chinchilla}.
All non-optimizer parts of the protocol (tokenizer, data splits,
checkpoint grid, seeds, and reporting rules) are fixed before the runs.

\begin{table}[t]
\centering
\caption{Experimental protocol.}
\label{tab:protocol}
\small
\begin{tabular}{p{0.33\linewidth}p{0.57\linewidth}}
\hline
Parameter & Value \\
\hline
Model configuration & Mamba-2 130M \\
Datasets & OpenWebText (OWT)~\cite{gokaslan2019openwebtext};
    FineWeb-Edu (FW)~\cite{penedo2024fineweb} \\
Token budgets & \(10^9\); \(2.6\times10^9\); \(5\times10^{10}\) \\
Batch / seq.\ len.\ / grad.\ accum.\ &
    \(8\) / \(2048\) / \(32\) (\(524{,}288\) tokens per step) \\
Main seeds & \(2\) shared seeds \(\{42,1337\}\) for the AdamW baseline
    and each regime \(k\) \\
Pilot & per-optimizer LR sweeps, \(2\times10^8\) tokens,
    seeds \(\{42,1337\}\) \\
Assignment regimes & \(k\in\{0,1,2,3\}\) (\(\mathcal{M}^{(k)}\)) \\
\hline
\end{tabular}
\end{table}

\paragraph*{Checkpoint grid and reporting indices}
For a training-token budget \(T_{\max}\) we define the evaluation
checkpoint grid as the \(N_{\mathrm{ck}}+1=11\) evenly spaced points
\[
    \mathcal{T}_{\mathrm{ck}}
    =
    \left\{
        \frac{i}{N_{\mathrm{ck}}}\, T_{\max}
        :
        i = 0, 1, \dots, N_{\mathrm{ck}}
    \right\},
    \qquad
    N_{\mathrm{ck}} = 10,
\]
so that \(\mathcal{T}_{\mathrm{ck}}\) explicitly contains \(0\),
\(T_{\max}/2\), and \(T_{\max}\); the extended-budget runs use the same
grid rescaled to \(T_{\max}=2.6\times10^9\).
For regime \(k\), seed \(r\), and evaluation point
\(t\in\mathcal{T}_{\mathrm{ck}}\), let
\(\theta_t^{(k,r)}\) denote the parameters after processing \(t\)
training tokens; the metrics computed from these checkpoints are
defined below.

\paragraph*{Evaluation metrics}
\begin{itemize}

\item Pilot hyperparameter selection uses the AdamW and Muon
learning-rate sweeps. Their outputs serve only to select stable
optimizer-specific hyperparameters before the primary paired comparison.
Pilot runs are archived for reproducibility but are not included in the
final comparison tables.

\item Validation quality and convergence are reported through
validation loss and validation perplexity at
\(t\in\mathcal{T}_{\mathrm{ck}}\), including the final values
\[
    \mathcal{L}_{\mathrm{val}}(\theta_{T_{\max}}^{(k,r)})
    \qquad\text{and}\qquad
    \mathrm{PPL}_{\mathrm{val}}(\theta_{T_{\max}}^{(k,r)}).
\]
For each regime, these quantities are reported as
mean~\(\pm\)~std across the available seeds.

\item Paired seed-wise differences use the same seed identifiers in the
AdamW baseline and at each Muon regime. We
report paired differences in final validation loss relative to the
\(k=0\) baseline:
\[
    \Delta^{\mathcal{L},k}_{r}
    =
    \mathcal{L}_{\mathrm{val}}
    \bigl(
        \theta_{T_{\max}}^{(k,r)}
    \bigr)
    -
    \mathcal{L}_{\mathrm{val}}
    \bigl(
        \theta_{T_{\max}}^{(0,r)}
    \bigr).
\]
An analogous paired difference may also be reported for final
validation perplexity when useful for interpretation. For each
\(k=1,2\), the primary paired comparison reports mean~\(\pm\)~std for
\(\Delta^{\mathcal{L},k}_{r}\) across the shared seeds. Because the
number of seeds is small, paired seed-wise differences with mean/std
are treated as descriptive summaries rather than formal significance
tests. For reporting purposes, Muon is considered to improve over AdamW at
regime \(k\) if
\[
    \frac{1}{R}
    \sum_{r=1}^{R}
    \Delta^{\mathcal{L},k}_{r}
    <
    0
\]
over the \(R\) completed stable shared seeds. A positive mean paired
difference is reported as worse than AdamW, while a near-zero mean is
reported as a null result under the fixed protocol.

\item Token efficiency is quantified by \emph{validation-equivalent
tokens}: for each AdamW checkpoint with mean validation loss \(\ell\), we
report the number of tokens each Muon regime needs to first reach
\(\ell\), obtained by linear interpolation of its validation-loss curve
over \(\mathcal{T}_{\mathrm{ck}}\). A value below the AdamW reference
(the diagonal) means Muon attains that loss in fewer tokens; this is
reported on OpenWebText at \(T_{\max}=10^9\) (Section~\ref{sec:results}).
The cross-budget narrowing of the baseline-to-\(k=3\) gap
(Section~\ref{sec:results}) is a complementary view of the same effect.

\item Ablation reporting follows a fixed suite of optimizer-modification
slices, each anchored at the primary configuration selected after the
primary paired comparison (the regime with the lowest mean final
validation loss) and fixed before any ablation runs. The slices are
RMS-style rescaling of the Muon update, explicit Muon-side weight decay,
and per-slice assignment of Muon within the merged input projection
(Appendix~\ref{sec:rms-wd-ablations}). They are reported separately from
the primary paired comparison and do not enter the seed-paired
\(\Delta^{\mathcal{L},k}_{r}\) summaries.

\item Secondary matrix diagnostics are computed on the representative
seed (seed~\(42\), the first of the two shared main seeds) and tracked
over training, which yields the trajectories in
Section~\ref{sec:spectral}. The layer-averaged summary is reported at
\(T_{\max}\) and the per-layer profiles at the subset
\[
    \mathcal{T}_{\mathrm{spec}}
    =
    \{T_{\max}/2, T_{\max}\}
    \subseteq
    \mathcal{T}_{\mathrm{ck}}.
\]
These diagnostics are used only to interpret
whether optimizer-dependent differences are associated with changes
in matrix spectra; they are not treated as primary success metrics or
as formal mechanistic proof.

Following Roy and Vetterli \cite{roy2007effective}, for a nonzero
matrix \(W\in\mathbb{R}^{d_1\times d_2}\), let
\(m=\min\{d_1,d_2\}\) and let
\(\sigma_1\ge\dots\ge\sigma_m\ge0\) denote its singular values. We
define
\[
    \bar{\sigma}_i
    =
    \frac{\sigma_i}{\sum_{j=1}^{m}\sigma_j},
\]
\[
    H(W)
    =
    -
    \sum_{i=1}^{m}
    \bar{\sigma}_i \log \bar{\sigma}_i,
    \qquad
    \mathrm{erank}(W)
    =
    \exp\!\bigl(H(W)\bigr),
\]
and
\[
    \kappa(W)
    =
    \frac{\sigma_1}{\sigma_m},
    \qquad
    \sigma_m>0.
\]
We use the standard entropy convention \(0\log 0=0\). If
\(\sigma_m=0\) or is numerically zero, the matrix is reported as
singular or ill-conditioned rather than assigned an arbitrary finite
condition number. Per-layer profiles and layer-averaged summaries of
spectral norm, Frobenius norm, effective rank, and condition number
are reported whenever the corresponding decomposition is numerically
well-defined. Zero matrices are skipped for effective-rank reporting.

\end{itemize}

\section{Full-budget training results}
\label{sec:results}

\subsection{Setup}

Using the learning rates of Table~\ref{tab:lr-summary}, we trained all
four regimes $k \in \{0,1,2,3\}$ over the two corpora (OWT, FW) and the
two token budgets (\(10^{9}\) and \(2.6\times 10^{9}\), abbreviated 1B
and 2.6B) of the protocol, with all non-optimizer hyperparameters held
identical across regimes, corpora, and budgets
(Section~\ref{sec:metrics}).
Section~\ref{ssec:owt-1b} covers the primary paired comparison (OWT at
1B), and Section~\ref{ssec:robustness} reports robustness across corpus
and budget.

\subsection{Final validation loss and perplexity on OpenWebText}
\label{ssec:owt-1b}

Table~\ref{tab:full-results} reports the final validation loss and
perplexity at $T_{\max}=10^{9}$ tokens on OpenWebText.
All three Muon regimes reduce final validation loss relative to the
AdamW baseline.
Of the two intermediate regimes, $k=2$
($\mathcal{G}_{\mathrm{in}}\cup\mathcal{G}_{\mathrm{out}}$) yields the
larger reduction ($\Delta\mathcal{L} = -0.071$, PPL $28.42$, a ${\approx}6.8\%$
reduction in perplexity) and $k=1$ ($\mathcal{G}_{\mathrm{in}}$) the
smaller ($\Delta\mathcal{L} = -0.031$, PPL $29.58$, ${\approx}3.0\%$).
Seed-wise variance is lower under Muon than under AdamW
($\pm 0.001$--$0.002$ versus $\pm 0.010$).
The $k=3$ regime ($\mathcal{G}_{\mathrm{out}}$ only) achieves the lowest
final loss of the four regimes ($\Delta\mathcal{L} = -0.116$, PPL $27.17$); this
result is discussed in Section~\ref{sec:discussion}.

\begin{table}[t]
  \centering
  \caption{Final validation loss and perplexity at
    $T_{\max}=10^{9}$ tokens on OpenWebText. Mean and standard deviation
    are computed over seeds $\{42,1337\}$. $\Delta\mathcal{L}$ is relative to
    $k=0$.}
  \label{tab:full-results}
  \scriptsize
  \begin{tabular}{c l c c c c}
    \hline
    $k$ & Muon groups & $\eta^{\star}_{\mathrm{Muon}}$ &
      $\mathcal{L}_{\mathrm{val}}$ & PPL & $\Delta\mathcal{L}$ \\
    \hline
    0 & AdamW & --- &
      $3.4178 \pm 0.0103$ & 30.50 & --- \\
    1 & $\mathcal{G}_{\mathrm{in}}$ & $10^{-2}$ &
      $3.3869 \pm 0.0020$ & 29.58 & $-0.031$ \\
    2 & $\mathcal{G}_{\mathrm{in}}\cup\mathcal{G}_{\mathrm{out}}$ &
      $2\times 10^{-2}$ &
      $3.3472 \pm 0.0012$ & 28.42 & $-0.071$ \\
    3 & $\mathcal{G}_{\mathrm{out}}$ &
      $1.5\times 10^{-2}$ &
      $3.3019 \pm 0.0025$ & 27.17 & $-0.116$ \\
    \hline
  \end{tabular}
\end{table}

\begin{figure}[t]
  \centering
  \includegraphics[width=\linewidth]{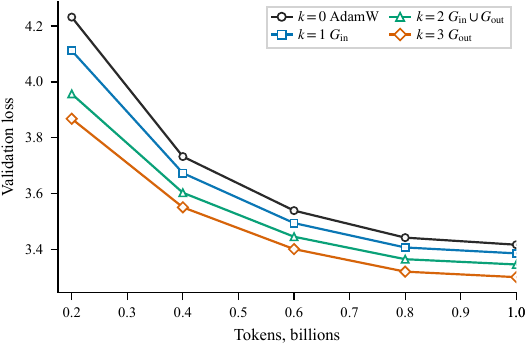}
  \caption{Validation loss at checkpoints $\mathcal{T}_{\mathrm{ck}}$,
    averaged over seeds $\{42,1337\}$ (OpenWebText, $T_{\max}=10^{9}$).}
  \label{fig:owt-val}
\end{figure}

\begin{figure}[t]
  \centering
  \includegraphics[width=\linewidth]{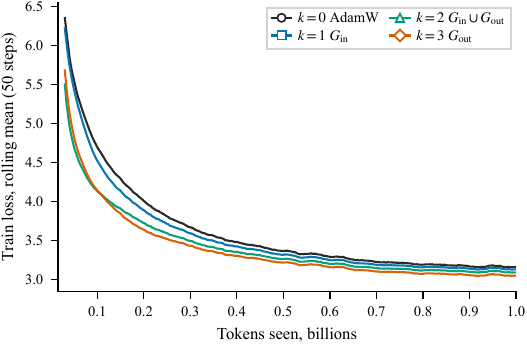}
  \caption{Smoothed training loss (rolling mean over 50 steps),
    averaged over seeds $\{42,1337\}$ (OpenWebText, $T_{\max}=10^{9}$).
    Curves begin at $0.03\times 10^{9}$ tokens.}
  \label{fig:owt-train}
\end{figure}

\begin{figure}[t]
  \centering
  \includegraphics[width=\linewidth]{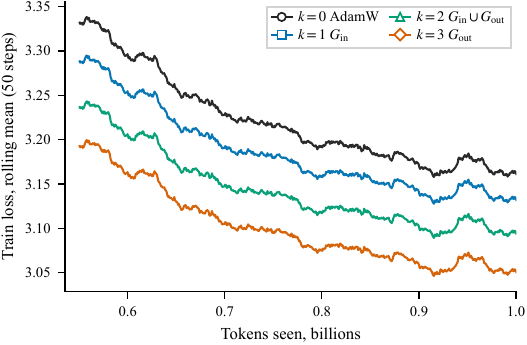}
  \caption{Smoothed training loss over the late training phase
    ($T \in [0.55, 1.0]\times 10^{9}$ tokens, OpenWebText).}
  \label{fig:owt-late}
\end{figure}

Figures~\ref{fig:owt-val} and~\ref{fig:owt-train} trace the validation
and training loss of this run.
The ranking $k=3 < k=2 < k=1 < k=0$ emerges by
$T = 2\times 10^{8}$ tokens, and the curves never cross through
$T_{\max}$. The late-phase training loss (Figure~\ref{fig:owt-late})
shows that the gaps persist under learning-rate decay, so they are not
merely an early-training transient.

\begin{figure}[t]
  \centering
  \includegraphics[width=\linewidth]{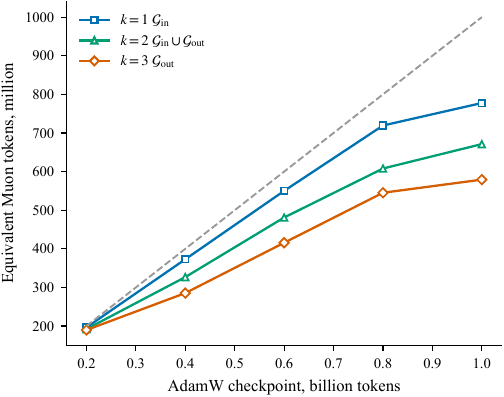}
  \caption{Validation-equivalent Muon tokens for matched AdamW loss
    (OpenWebText, seeds $\{42,1337\}$). For each AdamW checkpoint
    (horizontal axis), the vertical axis is the number of tokens each
    Muon regime needs to reach the same validation loss; the dashed
    diagonal is the AdamW reference. All regimes lie below it throughout
    training.}
  \label{fig:token-equiv}
\end{figure}

Beyond the final loss, Muon is more token-efficient at matched loss: for
every AdamW checkpoint, each Muon regime reaches the same validation loss
in fewer tokens, so all three curves lie below the AdamW reference
diagonal (Figure~\ref{fig:token-equiv}). The effect is largest for $k=3$,
which matches AdamW's final $10^{9}$-token validation loss in only
${\approx}5.8\times 10^{8}$ tokens.

\subsection{Robustness across corpus and training budget}
\label{ssec:robustness}

To test whether the OpenWebText ranking reflects the optimizer rather
than a single corpus or budget, we repeated the four-regime comparison
on FineWeb-Edu and at the extended $2.6\times 10^{9}$-token budget on
both corpora.
Table~\ref{tab:robustness} reports the final validation loss in all four
corpus--budget cells.

\begin{table}[t]
  \centering
  \caption{Mean final validation loss across corpus and training budget,
    over seeds $\{42,1337\}$. The bottom row is the gap between the AdamW
    baseline and the best regime, $k=3$.}
  \label{tab:robustness}
  \scriptsize
  \begin{tabular}{c l c c c c}
    \hline
    $k$ & Muon groups &
      OWT $1\mathrm{B}$ & OWT $2.6\mathrm{B}$ &
      FW $1\mathrm{B}$ & FW $2.6\mathrm{B}$ \\
    \hline
    0 & AdamW &
      $3.4178$ & $3.1813$ & $3.3589$ & $3.1508$ \\
    1 & $\mathcal{G}_{\mathrm{in}}$ &
      $3.3869$ & $3.1539$ & $3.3409$ & $3.1397$ \\
    2 & $\mathcal{G}_{\mathrm{in}}\cup\mathcal{G}_{\mathrm{out}}$ &
      $3.3472$ & $3.1528$ & $3.3246$ & $3.1457$ \\
    3 & $\mathcal{G}_{\mathrm{out}}$ &
      $3.3019$ & $3.1184$ & $3.2761$ & $3.1080$ \\
    \hline
    \multicolumn{2}{l}{$\Delta\mathcal{L}\,(k{=}0 - k{=}3)$} &
      $+0.116$ & $+0.063$ & $+0.083$ & $+0.043$ \\
    \hline
  \end{tabular}
\end{table}

Two findings hold in every cell.
First, the endpoints of the ranking are stable: $k=3$
($\mathcal{G}_{\mathrm{out}}$ only) has the lowest validation loss
and AdamW the highest, on both corpora and at both budgets.
Second, the advantage is mostly a token-efficiency effect. The
baseline-to-$k=3$ gap shrinks as the budget grows (from $0.116$ to
$0.063$ on OpenWebText, and from $0.083$ to $0.043$ on FineWeb-Edu),
so AdamW partially closes the gap with more tokens but does not overtake
$k=3$ within the budgets tested.
The intermediate ordering of $k=1$ and $k=2$ is not stable at
$2.6\times 10^{9}$ tokens: on OpenWebText the two coincide within noise
($3.1539$ versus $3.1528$), and on FineWeb-Edu $k=1$ overtakes $k=2$.
This matches the pilot observation that assigning both projection groups
to Muon ($k=2$) does not compound the single-group gains.

\begin{figure}[t]
  \centering
  \includegraphics[width=\linewidth]{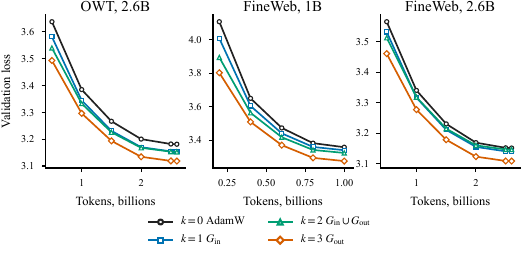}
  \caption{Validation loss at checkpoints $\mathcal{T}_{\mathrm{ck}}$ for
    the three remaining corpus--budget cells, averaged over seeds
    $\{42,1337\}$ (the primary OpenWebText $10^{9}$-token run is
    Figure~\ref{fig:owt-val}).}
  \label{fig:full-val}
\end{figure}

\begin{figure}[t]
  \centering
  \includegraphics[width=\linewidth]{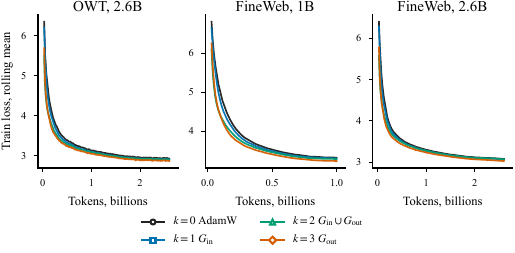}
  \caption{Smoothed training loss (rolling mean over 50 steps) for the
    three remaining corpus--budget cells, averaged over seeds
    $\{42,1337\}$.}
  \label{fig:full-train}
\end{figure}

\begin{figure}[t]
  \centering
  \includegraphics[width=\linewidth]{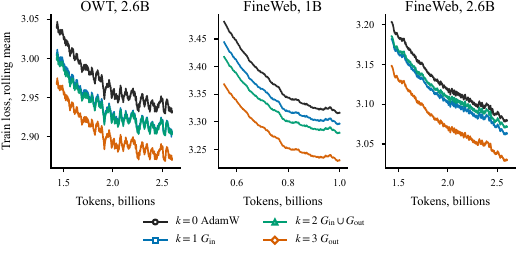}
  \caption{Smoothed training loss over the late training phase
    (final ${\sim}45\%$ of each budget) for the three remaining
    corpus--budget cells, averaged over seeds $\{42,1337\}$.}
  \label{fig:full-late}
\end{figure}

The loss curves show the same pattern in every remaining cell
(Figures~\ref{fig:full-val}--\ref{fig:full-late}): the $k=0$ and $k=3$
curves separate early and never cross, and only the intermediate $k=1$
and $k=2$ curves draw together at the extended budget.

\section{Spectral diagnostics}
\label{sec:spectral}

\begin{figure}[t]
  \centering
  \includegraphics[width=\linewidth]{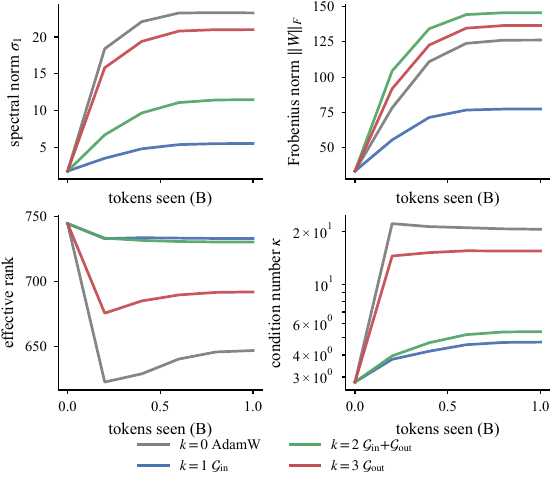}
  \caption{Layer-averaged spectral metrics for $\mathcal{G}_{\mathrm{in}}$
    over training, seed~$42$. The four panels show spectral norm
    $\sigma_1$, Frobenius norm, effective rank, and condition number
    $\kappa$.}
  \label{fig:spectral-traj-gin}
\end{figure}

\begin{figure}[t]
  \centering
  \includegraphics[width=\linewidth]{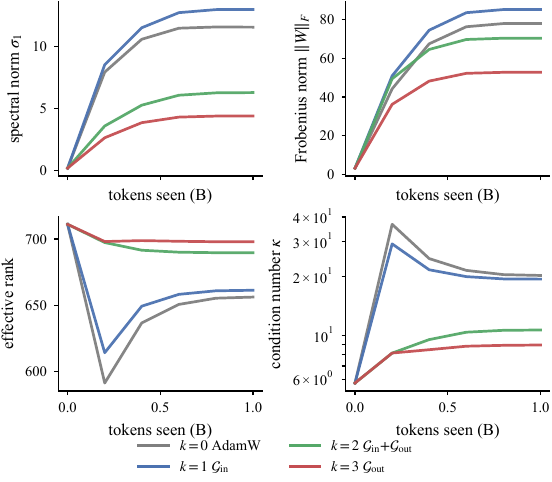}
  \caption{Layer-averaged spectral metrics for $\mathcal{G}_{\mathrm{out}}$
    over training, seed~$42$. Panels as in
    Figure~\ref{fig:spectral-traj-gin}.}
  \label{fig:spectral-traj-gout}
\end{figure}

Spectral diagnostics were computed on seed~$42$ and tracked over
training. Table~\ref{tab:spectral} reports layer-averaged values at
$T_{\max}$; Figures~\ref{fig:spectral-traj-gin}
and~\ref{fig:spectral-traj-gout} show how these quantities evolve over
training. Per-layer profiles
at $T_{\max}$ and the replication across corpus--budget cells are
deferred to Appendix~\ref{sec:spectral-detail}.
These analyses are interpretive, not primary success metrics.

\begin{table}[t]
  \centering
  \caption{Layer-averaged spectral diagnostics at $T_{\max}$, seed~$42$.
    Effective rank and condition number $\kappa$ are defined in
    Section~\ref{sec:metrics}.}
  \label{tab:spectral}
  \footnotesize
  \setlength{\tabcolsep}{3.5pt}
  \begin{tabular}{llrrr}
    \hline
    Regime & Group & $\sigma_1$ & erank & $\kappa$ \\
    \hline
    $k=0$ AdamW &
      $\mathcal{G}_{\mathrm{in}}$  & 23.29 & 646.7 & 20.56 \\
    $k=1$ $\mathcal{G}_{\mathrm{in}}$ &
      $\mathcal{G}_{\mathrm{in}}$  &  5.53 & 732.9 &  4.73 \\
    $k=2$ $\mathcal{G}_{\mathrm{in}}\cup\mathcal{G}_{\mathrm{out}}$ &
      $\mathcal{G}_{\mathrm{in}}$  & 11.49 & 730.2 &  5.42 \\
    $k=3$ $\mathcal{G}_{\mathrm{out}}$ &
      $\mathcal{G}_{\mathrm{in}}$  & 21.01 & 691.8 & 15.47 \\
    \hline
    $k=0$ AdamW &
      $\mathcal{G}_{\mathrm{out}}$ & 11.56 & 656.1 & 20.18 \\
    $k=1$ $\mathcal{G}_{\mathrm{in}}$ &
      $\mathcal{G}_{\mathrm{out}}$ & 12.98 & 661.3 & 19.33 \\
    $k=2$ $\mathcal{G}_{\mathrm{in}}\cup\mathcal{G}_{\mathrm{out}}$ &
      $\mathcal{G}_{\mathrm{out}}$ &  6.29 & 689.6 & 10.66 \\
    $k=3$ $\mathcal{G}_{\mathrm{out}}$ &
      $\mathcal{G}_{\mathrm{out}}$ &  4.40 & 697.9 &  8.95 \\
    \hline
  \end{tabular}
\end{table}

The spectral effect of Muon is confined to the group to which it is
assigned.
For $\mathcal{G}_{\mathrm{in}}$, the $k=1$ regime reduces $\kappa$
from $20.56$ to $4.73$ and increases effective rank from $646.7$ to
$732.9$, while the $\mathcal{G}_{\mathrm{out}}$ condition number in
the same regime remains at $19.33$, nearly unchanged from the AdamW
baseline.
Conversely, $k=3$ achieves $\kappa = 8.95$ and erank $= 697.9$ on
$\mathcal{G}_{\mathrm{out}}$, while the $\mathcal{G}_{\mathrm{in}}$
condition number under $k=3$ remains at $15.47$.
The joint assignment $k=2$ improves both groups moderately but reaches
neither the $\mathcal{G}_{\mathrm{in}}$ conditioning of $k=1$ nor
the $\mathcal{G}_{\mathrm{out}}$ conditioning of $k=3$.

The spectral separation between regimes emerges by
$T = 2\times 10^{8}$ tokens and changes little thereafter
(Figures~\ref{fig:spectral-traj-gin},~\ref{fig:spectral-traj-gout}).
For $\mathcal{G}_{\mathrm{in}}$, the condition number under $k=1$
rises to ${\approx}4$ by $0.2\times 10^{9}$ tokens and plateaus,
while the AdamW baseline reaches ${\approx}20$ and stabilizes there.
For $\mathcal{G}_{\mathrm{out}}$, the effective rank under $k=0$ and
$k=1$ drops sharply at $0.2\times 10^{9}$ tokens before partially
recovering; $k=3$ avoids this drop and maintains the highest effective
rank throughout training.
The per-layer profiles confirm that this structure is consistent across
all 24 layers with no systematic depth dependence
(Appendix~\ref{sec:spectral-detail}).
Taken together, the spectral evidence supports the reading that
well-conditioned output projections are the main source of Muon's
empirical advantage here.

\section{Does the advantage survive at scale?}
\label{ssec:durability}

Within the budgets of Section~\ref{ssec:robustness} the baseline-to-$k=3$ gap
shrinks monotonically with tokens, raising the question of whether the
$\mathcal{G}_{\mathrm{out}}$ advantage is only an early-training effect. To
test this we trained the two endpoint regimes, $k=0$ (AdamW) and $k=3$ (Muon
on $\mathcal{G}_{\mathrm{out}}$), far past the compute-optimal point:
Mamba-2 130M on FineWeb-Edu to $5\times 10^{10}$ tokens
($\approx 19\times$ the Chinchilla-optimal budget, deeply overtrained),
single seed, all non-optimizer hyperparameters held fixed.

\begin{figure*}[t]
  \centering
  \includegraphics[width=\linewidth]{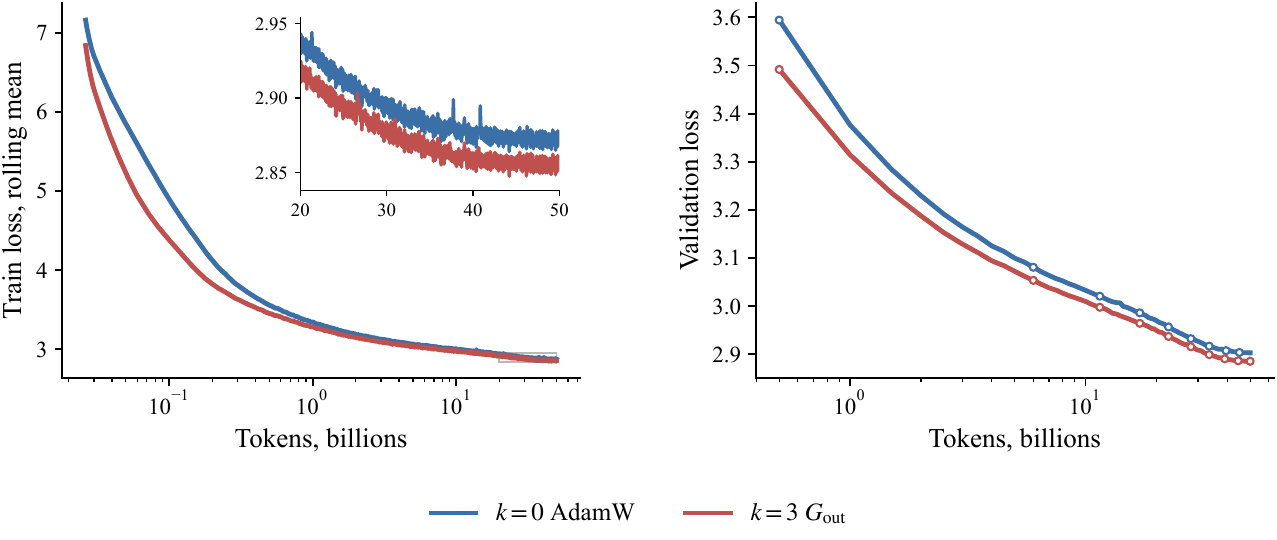}
  \caption{Training (left) and validation (right) loss for $k=0$ and $k=3$,
    Mamba-2 130M on FineWeb-Edu to $5\times 10^{10}$ tokens
    ($\approx 19\times$ Chinchilla), single seed. $k=3$ stays below the AdamW
    baseline throughout; final validation loss $2.8853$ vs.\ $2.9027$.}
  \label{fig:durability}
\end{figure*}

The $k=3$ advantage persists. The Muon-$\mathcal{G}_{\mathrm{out}}$ curve
stays below the AdamW baseline on both train and validation at every
evaluation checkpoint, ending at a final validation loss of $2.8853$
versus $2.9027$ (a gap of $0.017$; Figure~\ref{fig:durability}). The gap
continues to narrow with budget, consistent with the token-efficiency
reading of Section~\ref{ssec:owt-1b}, but AdamW does not overtake $k=3$
even at $19\times$ Chinchilla. The output-projection assignment is therefore a
durable advantage in pre-training loss, not an artifact of under-budgeted
runs.

\section{Downstream evaluation}
\label{ssec:downstream}

A lower pre-training loss is only useful if it transfers to downstream tasks.
We evaluated the two 50B checkpoints of Section~\ref{ssec:durability} with
\texttt{lm-evaluation-harness} in the zero-shot setting, following the
evaluation protocol of Mamba and Mamba-2~\cite{gu2023mamba,dao2024mamba2} on
seven standard tasks (Table~\ref{tab:downstream}).

The $0.017$ validation-loss gap does \emph{not} translate into a downstream
advantage. Neither regime is consistently better: Muon wins four of the
seven tasks and AdamW three, and the mean accuracy differs by only $0.3$
points ($41.7$ vs.\ $42.0$). The per-task gaps are small and mostly within
noise; the clearest exception is LAMBADA, where AdamW leads on both
accuracy and perplexity. At 130M, pre-training loss and zero-shot accuracy
decouple. We read this not as a weakness of the output-projection
assignment but as a reminder that small-model loss gaps need not carry to
task accuracy, consistent with the near-chance behavior of models of this
size on several of these tasks in prior work~\cite{gu2023mamba,dao2024mamba2}.

\begin{table}[t]
  \centering
  \caption{Zero-shot accuracy (\%) at $5\times 10^{10}$ tokens
    (\textsc{lm-evaluation-harness}, 0-shot; Mamba-2 130M, FineWeb-Edu, seed
    42). Per-task winner underlined; $\pm$ standard error. acc$_n$ denotes
    normalized accuracy. Bottom rows: mean accuracy and LAMBADA perplexity.}
  \label{tab:downstream}
  \scriptsize
  \begin{tabular}{l l c c}
    \hline
    Task & Metric & AdamW ($k{=}0$) & Muon $\mathcal{G}_{\mathrm{out}}$ ($k{=}3$) \\
    \hline
    LAMBADA       & acc      & \underline{27.6}\,{\tiny$\pm$0.6} & 25.5\,{\tiny$\pm$0.6} \\
    HellaSwag     & acc$_n$  & 35.7\,{\tiny$\pm$0.5} & \underline{36.6}\,{\tiny$\pm$0.5} \\
    PIQA          & acc      & \underline{66.2}\,{\tiny$\pm$1.1} & 64.1\,{\tiny$\pm$1.1} \\
    ARC-Easy      & acc      & 56.2\,{\tiny$\pm$1.0} & \underline{58.5}\,{\tiny$\pm$1.0} \\
    ARC-Challenge & acc$_n$  & \underline{26.1}\,{\tiny$\pm$1.3} & 25.5\,{\tiny$\pm$1.3} \\
    WinoGrande    & acc      & 49.3\,{\tiny$\pm$1.4} & \underline{51.6}\,{\tiny$\pm$1.4} \\
    OpenBookQA    & acc$_n$  & 30.6\,{\tiny$\pm$2.1} & \underline{31.8}\,{\tiny$\pm$2.1} \\
    \hline
    \textbf{Average} & acc & 41.7 & \underline{42.0} \\
    LAMBADA       & ppl\,$\downarrow$ & \underline{63.3} & 68.5 \\
    \hline
  \end{tabular}
\end{table}
\section{Discussion}
\label{sec:discussion}

Across our four assignment regimes, Muon on \texttt{out\_proj} alone
($k=3$) consistently outperforms the joint assignment ($k=2$) and
the \texttt{in\_proj}-only assignment ($k=1$). The ordering stabilizes
by roughly $2\times 10^{8}$ tokens, and seed-wise variation is an
order of magnitude smaller than the inter-regime gaps. One
interpretation, consistent with the architecture of Mamba-2, is that
\texttt{out\_proj} is a uniform dense projection, whereas
\texttt{in\_proj} is a row-concatenation of five functionally distinct
sub-projections ($z$, $x$, $B$, $C$, $\Delta t$). Muon's
Newton--Schulz iteration orthogonalizes the update spectrum across
the full matrix, a global operation that may be ill-matched to a
stacked projection whose sub-blocks need different update geometries.

The gain is concentrated in early-to-mid training rather than in a lower
loss floor. The baseline-to-$k=3$ gap shrinks monotonically with the
budget (from $0.116$ to $0.063$ on OpenWebText and from $0.083$ to
$0.043$ on FineWeb-Edu between $10^9$ and $2.6\times10^9$ tokens), and
$k=3$ reaches AdamW's $10^9$-token loss in roughly $5.8\times10^8$
tokens. We read this as a token-efficiency effect: Muon on
$\mathcal{G}_{\mathrm{out}}$ buys a given loss in fewer tokens rather
than a different asymptote, so its practical value is largest in
compute-constrained, near-single-epoch training and should fade as
training approaches convergence.

The spectral diagnostics in Section~\ref{sec:spectral} provide an
independent line of evidence consistent with this reading. Muon's
spectral effect is assignment-local: the condition number of each
projection group changes only when Muon is assigned to that group.
The $\mathcal{G}_{\mathrm{out}}$ condition number decreases
monotonically with the strength of Muon's involvement
(AdamW $20.18 \to k=1\ 19.33 \to k=2\ 10.66 \to k=3\ 8.95$), and the
regime that achieves the cleanest $\mathcal{G}_{\mathrm{out}}$
spectrum ($k=3$) is also the one with the lowest validation loss.
Conversely, the largest improvement in $\mathcal{G}_{\mathrm{in}}$
conditioning at $k=1$, where $\kappa$ drops from $20.56$ to $4.73$,
is not matched by a comparable gain in validation loss; a
better-conditioned \texttt{in\_proj} is not on its own sufficient.
We do not treat this as causal evidence, since the spectral
diagnostics use a single seed, but the loss ranking and the spectral
conditioning point to the same projection group.

Two follow-up analyses sharpen this reading rather than just restating
it. Resolving \texttt{in\_proj} into its five row-slices
(Appendix~\ref{sec:spectral-detail}) shows that Muon conditions the
ill-posed slices (the gating channel $z$ and the data channel $x$, whose
AdamW condition numbers are $\kappa\approx25$--$40$) while leaving the
already-isotropic $B$, $C$, and $\Delta t$ essentially unchanged. Yet
assigning Muon to those slices on top of $k=3$
(Appendix~\ref{ssec:subblocks}) does not lower the loss: the gating
channel $z$ is the single most harmful slice, and the data channels
$x,B,C$ help only at the $2\times10^8$-token pilot, reversing at
$2.6\times10^9$ tokens. Conditioning of \texttt{in\_proj}, whole or
per-slice, is therefore \emph{decoupled} from loss, and the benefit is
specific to the homogeneous output projection rather than to
conditioning the input projection at all.

Two caveats temper the mechanism. The Muon learning rates come from a
discrete two-seed grid, so we cannot fully exclude that $k=2$'s
under-performance reflects a single scalar rate serving heterogeneous
sub-blocks; and the spectra use a single seed. Our
data also do not cleanly separate \texttt{out\_proj}'s role as the
residual-stream writer from \texttt{in\_proj}'s heterogeneity per se.
Even so, three independent signals (the loss ranking, the group-level
spectra, and the per-slice analysis) point to the same conclusion: among
Mamba-2's weight groups, Muon helps where the matrix is homogeneous and
writes to the residual stream.

\appendices
\section{Learning-rate calibration}
\label{sec:lr-calibration}

\subsection{AdamW learning-rate calibration}

The AdamW baseline learning rate was calibrated under the fixed pilot
protocol described in Section~\ref{sec:metrics}.
Each run used Mamba-2 130M on OpenWebText with micro-batch size~$8$,
sequence length~$2048$, gradient accumulation~$32$, and pilot budget
$T = 2\times 10^{8}$ training tokens, so each optimizer step processed
\[
  8 \times 2048 \times 32 = 524{,}288 \text{ tokens.}
\]
The only varied hyperparameter was the AdamW learning rate $\alpha$.

\begin{figure}[t]
  \centering
  \includegraphics[width=\linewidth]{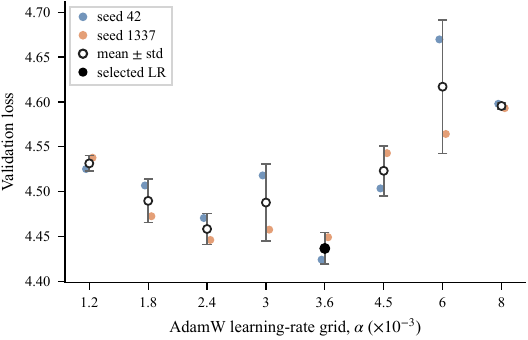}
  \caption{AdamW learning-rate sweep at the pilot budget
    $T = 2\times 10^{8}$ training tokens.
    Colored markers show individual seeds $\{42, 1337\}$; open markers
    with error bars show mean~$\pm$~standard deviation.
    The filled marker indicates the selected learning rate,
    $\alpha^{\star}_{\mathrm{AdamW}} = 3.6\times 10^{-3}$.}
  \label{fig:adamw-lr-sweep}
\end{figure}

Figure~\ref{fig:adamw-lr-sweep} shows the sweep over
\[
  \alpha \in
  \{1.2,\;1.8,\;2.4,\;3.0,\;3.6,\;4.5,\;6.0,\;8.0\}
  \times 10^{-3},
\]
evaluated on the shared pilot seeds $\{42, 1337\}$.
Colored markers show per-seed results; open markers with error bars show
the two-seed mean and standard deviation; the filled marker indicates
the selected value.
The minimum two-seed mean validation loss was attained at
$\alpha^{\star}_{\mathrm{AdamW}} = 3.6\times 10^{-3}$,
with per-seed values $4.4243$ and $4.4493$, giving
$\mathcal{L}_{\mathrm{val}} = 4.4368 \pm 0.0177$.
This value was frozen for all subsequent runs.

\subsection{Muon learning-rate calibration}

After fixing the AdamW learning rate, the Muon learning rate was
calibrated separately for each regime $k \in \{1,2,3\}$.
In all Muon pilot runs the AdamW learning rate for non-Muon parameters
was held at $\alpha^{\star}_{\mathrm{AdamW}} = 3.6\times 10^{-3}$;
all other training constants were identical to the AdamW pilot protocol.

\begin{figure}[t]
  \centering
  \includegraphics[width=\linewidth]{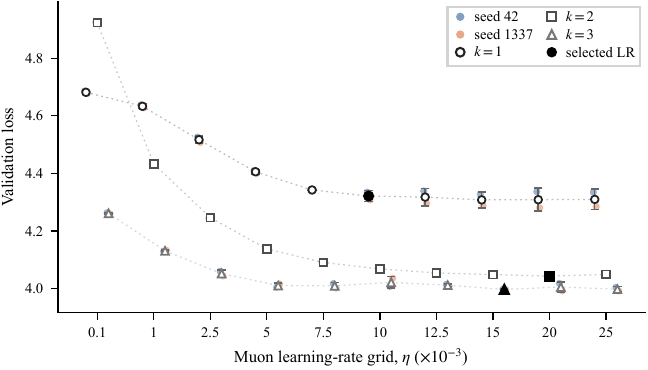}
  \caption{Muon learning-rate sweep at the pilot budget
    $T = 2\times 10^{8}$ training tokens, with AdamW learning rate
    fixed to $\alpha^{\star}_{\mathrm{AdamW}} = 3.6\times 10^{-3}$
    for all non-Muon parameters.
    Colored markers show individual seeds $\{42, 1337\}$; open markers
    with error bars show mean~$\pm$~standard deviation.
    Filled markers indicate the selected Muon learning rate for each
    assignment regime.}
  \label{fig:muon-lr-sweep}
\end{figure}

The Muon learning-rate grid was
\begin{multline*}
  \eta \in
  \bigl\{
    10^{-4},\;10^{-3},\;2.5\times 10^{-3},\;5\times 10^{-3},\;
    7.5\times 10^{-3},\;10^{-2}, \\
    1.25\times 10^{-2},\;1.5\times 10^{-2},\;
    2\times 10^{-2},\;2.5\times 10^{-2}
  \bigr\},
\end{multline*}
evaluated on seeds $\{42, 1337\}$ for each regime.
Figure~\ref{fig:muon-lr-sweep} reports the resulting validation losses.
For each regime the selected learning rate minimized the two-seed mean
validation loss, subject to the requirement that both seeds complete the
pilot budget without divergence or non-finite loss values.
The one exception is $k=1$, where the sweep exhibited a broad plateau
whose raw minimum is statistically indistinguishable from neighboring
values; there a more stable point within the plateau was adopted, as
detailed below.

The selected values and corresponding pilot validation losses are
\begin{align*}
  \eta^{\star}_{k=1} &= 1.0\times 10^{-2}, &
    \mathcal{L}_{\mathrm{val}} &= 4.3214 \pm 0.0192, \\
  \eta^{\star}_{k=2} &= 2.0\times 10^{-2}, &
    \mathcal{L}_{\mathrm{val}} &= 4.0433 \pm 0.0026, \\
  \eta^{\star}_{k=3} &= 1.5\times 10^{-2}, &
    \mathcal{L}_{\mathrm{val}} &= 3.9971 \pm 0.0003.
\end{align*}

For $k=1$, the sweep revealed a broad plateau over
$\eta \in [1.0, 2.5]\times 10^{-2}$, within which mean validation loss
varied by less than $0.015$.
At the plateau minimum ($\eta = 1.5\times 10^{-2}$) the seed-wise
standard deviation was $0.028$, comparable to the improvement over
$\eta = 1.0\times 10^{-2}$ ($\Delta\mathcal{L} = 0.013$); the two values are
therefore statistically indistinguishable at the pilot sample size.
Since $\eta = 1.0\times 10^{-2}$ lies within the reference range
$[5\times 10^{-3}, 2\times 10^{-2}]$ of Jordan~\cite{jordan2024muonblog}
and Liu et al.~\cite{liu2025muon}, it was adopted as the full-run
setting for $k=1$.
For $k=2$, the sweep decreased with low seed-wise variance to an
unambiguous minimum at $\eta = 2.0\times 10^{-2}$.

The $k=3$ regime (Muon on $\mathcal{G}_{\mathrm{out}}$ only,
defined in Section~\ref{sec:problem}) was calibrated in the same way.
At $\eta^{\star}_{k=3} = 1.5\times 10^{-2}$ its pilot loss was the
lowest across all regimes ($\mathcal{L}_{\mathrm{val}} = 3.9971 \pm 0.0003$).

All three selected learning rates were frozen before any full-budget
run was inspected.

\subsection{Learning-rate transfer check}
\label{sec:lr-fineweb}

Rather than repeating the full grid search on
FineWeb-Edu~\cite{penedo2024fineweb}, we ran a reduced one-seed
three-point check at the $2\times 10^{8}$-token pilot budget to verify
that the OWT-calibrated learning rates transfer to the second corpus.
For the AdamW baseline ($k=0$) we swept
$\alpha \in \{2.4,\;3.6,\;4.5\}\times 10^{-3}$, covering the
neighborhood of the OWT optimum; the minimum was again attained at
$\alpha^{\star} = 3.6\times 10^{-3}$.
For the $k=3$ regime we swept
$\eta \in \{1.0,\;1.5,\;2.2\}\times 10^{-2}$; the minimum was again at
$\eta^{\star} = 1.5\times 10^{-2}$.
Since both optima coincided with the OWT selections, we transferred the
full set of OWT learning rates to FineWeb-Edu without modification.
Table~\ref{tab:lr-summary} summarizes the per-regime learning rates used
in all full-budget runs.

\begin{table}[t]
  \centering
  \caption{Learning rates used in all full-budget runs.
    The AdamW rate $\alpha^{\star}_{\mathrm{AdamW}} = 3.6\times 10^{-3}$
    is shared across all regimes and both corpora.
    The Muon rates for $k=1$ and $k=2$ on FineWeb-Edu were not
    independently swept; they were transferred from the OWT calibration
    after the $k=0$ and $k=3$ checks above confirmed transfer.}
  \label{tab:lr-summary}
  \scriptsize
  \begin{tabular}{c l c c}
    \hline
    $k$ & Muon groups &
      $\alpha^{\star}_{\mathrm{AdamW}}$ &
      $\eta^{\star}_{\mathrm{Muon}}$ \\
    \hline
    0 & --- (AdamW only) &
      $3.6\times 10^{-3}$ & --- \\
    1 & $\mathcal{G}_{\mathrm{in}}$ &
      $3.6\times 10^{-3}$ & $1.0\times 10^{-2}$ \\
    2 & $\mathcal{G}_{\mathrm{in}}\cup\mathcal{G}_{\mathrm{out}}$ &
      $3.6\times 10^{-3}$ & $2.0\times 10^{-2}$ \\
    3 & $\mathcal{G}_{\mathrm{out}}$ &
      $3.6\times 10^{-3}$ & $1.5\times 10^{-2}$ \\
    \hline
  \end{tabular}
\end{table}
\section{Additional analyses}
\label{sec:rms-wd-ablations}

This appendix collects analyses that support but did not feed into the
primary comparison: two ablations run after the learning-rate
calibration (RMS scaling and Muon weight decay) and an ablation that
resolves \texttt{in\_proj} into its individual sub-blocks. None of them
were used to select the primary regimes in Table~\ref{tab:full-results}.

\subsection{RMS scaling}

We tested whether RMS-style rescaling of the Muon update improves
over the calibrated non-RMS setting.
Short pilot runs were repeated with RMS scaling enabled for each
$k \in \{1,2,3\}$, with AdamW learning rate fixed at
$\alpha^{\star}_{\mathrm{AdamW}} = 3.6\times 10^{-3}$, Muon weight
decay set to zero, and the grid
\[
  \eta \in \{2.4\times 10^{-3},\;3.6\times 10^{-3},\;6.0\times 10^{-3}\}
\]
evaluated on seeds $\{42,1337\}$.
The best RMS-scaled pilot losses were
\[
\begin{aligned}
  k=1:\quad &4.3146 \pm 0.0440
    &&\text{at } \eta=2.4\times 10^{-3},\\
  k=2:\quad &4.0550 \pm 0.0049
    &&\text{at } \eta=2.4\times 10^{-3},\\
  k=3:\quad &4.0009 \pm 0.0042
    &&\text{at } \eta=3.6\times 10^{-3}.
\end{aligned}
\]
For $k=2$ and $k=3$ these are strictly worse than the corresponding
non-RMS minima ($4.0433 \pm 0.0026$ and $3.9971 \pm 0.0003$
from Section~\ref{sec:lr-calibration}).
For $k=1$ the RMS point improves the two-seed mean by $0.007$,
but the seed-wise standard deviation at that point is $0.044$,
so the difference is not reliable.
RMS scaling is therefore disabled in the main protocol.

\subsection{Weight decay for Muon}

We tested explicit Muon-side weight decay for $k=2$ and $k=3$,
using the Muon learning rates selected in
Section~\ref{sec:lr-calibration} ($\eta^{\star}_{k=2} = 2\times 10^{-2}$,
$\eta^{\star}_{k=3} = 1.5\times 10^{-2}$) and sweeping
\[
  \lambda_{\mathrm{Muon}} \in \{0.01,\;0.03,\;0.10\}
\]
on seeds $\{42,1337\}$.

For $k=2$ all three values lie within $0.002$ of the no-WD pilot
loss ($4.0433 \pm 0.0026$):
\[
\begin{aligned}
  \lambda=0.01 &:\quad 4.0418 \pm 0.0004,\\
  \lambda=0.03 &:\quad 4.0423 \pm 0.0017,\\
  \lambda=0.10 &:\quad 4.0418 \pm 0.0000.
\end{aligned}
\]

For $k=3$ the best result was obtained at $\lambda=0.03$:
\[
\begin{aligned}
  \lambda=0.01 &:\quad 4.0060 \pm 0.0017,\\
  \lambda=0.03 &:\quad 3.9926 \pm 0.0098,\\
  \lambda=0.10 &:\quad 4.0063 \pm 0.0030.
\end{aligned}
\]
This suggests a marginal benefit of moderate weight decay for the
$k=3$ regime, but the observation is based on the pilot
budget and was made after the primary full-budget runs were fixed;
it is treated as a follow-up hypothesis.

\subsection{Input-projection sub-blocks}
\label{ssec:subblocks}

\begin{figure}[t]
  \centering
  \includegraphics[width=\linewidth]{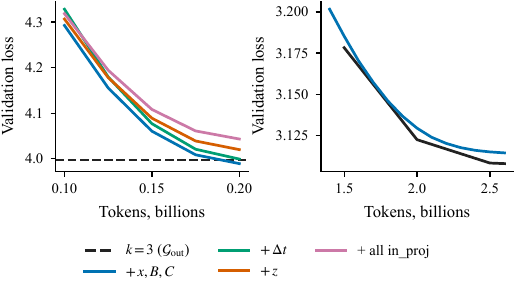}
  \caption{Validation loss for Muon on \texttt{in\_proj} sub-blocks added
    on top of $k=3$ (late-phase zoom). \emph{Left:} OWT pilot
    ($2\times10^{8}$ tokens); only $x,B,C$ falls below the $k=3$
    reference (dashed). \emph{Right:} FineWeb-Edu at $2.6\times10^{9}$
    tokens; the ordering inverts and $+\,x,B,C$ ends above $k=3$.}
  \label{fig:subblocks}
\end{figure}

\begin{figure}[t]
  \centering
  \includegraphics[width=\linewidth]{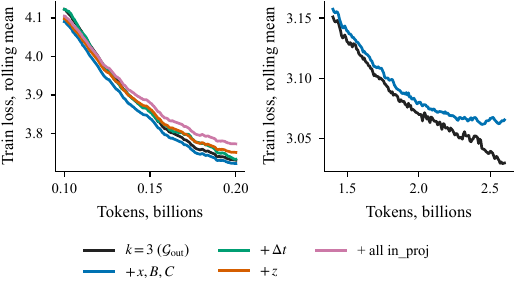}
  \caption{Smoothed training loss (rolling mean over 50 steps) for the
    sub-block ablation, late-phase zoom. \emph{Left:} OWT pilot
    ($2\times10^{8}$ tokens). \emph{Right:} FineWeb-Edu at
    $2.6\times10^{9}$ tokens. The ordering matches the validation loss:
    $x,B,C$ edges below $k=3$ on the pilot but ends above it at scale.}
  \label{fig:subblocks-train}
\end{figure}

The merged \texttt{in\_proj} weight is a row-concatenation of five
functionally distinct sub-blocks: the gating channel $z$, the data
channel $x$, the SSD observation projections $B,C$, and the time-step
$\Delta t$. A natural hypothesis (Section~\ref{sec:discussion}) is that
this heterogeneity is why Muon on $\mathcal{G}_{\mathrm{in}}$ does not
help, i.e.\ why $k=2$ trails $k=3$. We probe which sub-blocks help or
hurt by assigning Muon to individual row-slices \emph{on top of} $k=3$
(so $\mathcal{G}_{\mathrm{out}}$ is always under Muon), sweeping the Muon
learning rate per variant on the $2\times10^{8}$-token OWT pilot
(Table~\ref{tab:subblocks}; Figure~\ref{fig:subblocks}, left).
Only the data channels $x,B,C$ edge below the $k=3$ baseline
($\Delta\mathcal{L} = -0.0076$); the gating channel $z$ and the full
\texttt{in\_proj} ($=k{=}2$) are clearly worse. The five curves stay
cleanly ordered through the late phase and the same way on validation
and training loss (Figures~\ref{fig:subblocks} and~\ref{fig:subblocks-train}),
so this is a stable ranking rather than an end-point fluctuation. What
matters is which slices are added, not how many: the gating channel $z$
is the single most harmful slice and the full \texttt{in\_proj} ($k{=}2$)
the worst overall, which is what the heterogeneity hypothesis predicts.

This pilot advantage does not survive at scale. Re-running the best
candidate ($+\,x,B,C$) at the Chinchilla budget ($2.6\times10^{9}$
tokens) on FineWeb-Edu (four seeds, the hardest corpus--budget cell)
inverts the sign: $k=3$ reaches $3.1080$ while $+\,x,B,C$ reaches
$3.1145$ ($\Delta\mathcal{L} = +0.0065$, $\approx10\sigma$;
Figure~\ref{fig:subblocks}, right). The $200$M signal was therefore a
pilot-scale artifact, and $k=3$ (Muon on $\mathcal{G}_{\mathrm{out}}$
alone) remains the best configuration on every axis tested. The smoothed
training loss shows the same ordering in both regimes
(Figure~\ref{fig:subblocks-train}), and the $k=3$ margin over $+\,x,B,C$
widens with the budget rather than shrinking. In spectral terms this
mirrors Section~\ref{sec:spectral}: $k=1$ sharply improves the
conditioning of $\mathcal{G}_{\mathrm{in}}$ yet does not lower the loss,
so conditioning the input projections (whole or sliced) is not what
drives Muon's gain; the output projection $\mathcal{G}_{\mathrm{out}}$
is.

\begin{table}[t]
  \centering
  \caption{Input-projection sub-block ablation, with Muon added on top
    of $k=3$. Pilot: OWT at $2\times10^{8}$ tokens, best Muon rate per
    variant, mean over seeds $\{42,1337\}$. Validation: FineWeb-Edu at
    $2.6\times10^{9}$ tokens, four seeds. $\Delta\mathcal{L}$ is relative to
    $k=3$.}
  \label{tab:subblocks}
  \scriptsize
  \begin{tabular}{l c c}
    \hline
    Muon target & $\mathcal{L}_{\mathrm{val}}$ & $\Delta\mathcal{L}$ vs $k=3$ \\
    \hline
    \multicolumn{3}{l}{\emph{Pilot: OpenWebText @ $2\times10^{8}$ tokens}}\\
    $k=3$ ($\mathcal{G}_{\mathrm{out}}$) & $3.9971$ & --- \\
    $+\,x,B,C$ & $3.9895$ & $-0.0076$ \\
    $+\,\Delta t$ & $3.9995$ & $+0.0024$ \\
    $+\,z$ & $4.0202$ & $+0.0231$ \\
    $+\,$all \texttt{in\_proj} ($=k{=}2$) & $4.0435$ & $+0.0464$ \\
    \hline
    \multicolumn{3}{l}{\emph{Validation: FineWeb-Edu @ $2.6\times10^{9}$ tokens}}\\
    $k=3$ ($\mathcal{G}_{\mathrm{out}}$) & $3.1080$ & --- \\
    $+\,x,B,C$ & $3.1145$ & $+0.0065$ \\
    \hline
  \end{tabular}
\end{table}

In summary, none of these auxiliary modifications improves on the
calibrated plain $k=3$ configuration. RMS scaling does not improve
over the non-RMS settings. Muon weight decay has negligible effect on
$k=2$, with only a marginal $k=3$ benefit at moderate
$\lambda_{\mathrm{Muon}}$ that warrants a dedicated full-budget run.
Assigning Muon to individual \texttt{in\_proj} sub-blocks helps only at
the pilot scale, not at $2.6\times10^{9}$ tokens.                       
\section{Detailed spectral diagnostics}
\label{sec:spectral-detail}

This appendix collects the spectral diagnostics deferred from
Section~\ref{sec:spectral}: the per-layer profiles at $T_{\max}$, the
replication of the conditioning result across the remaining
corpus--budget cells, and the per-slice conditioning of \texttt{in\_proj}.

\begin{figure}[t]
  \centering
  \includegraphics[width=\linewidth]{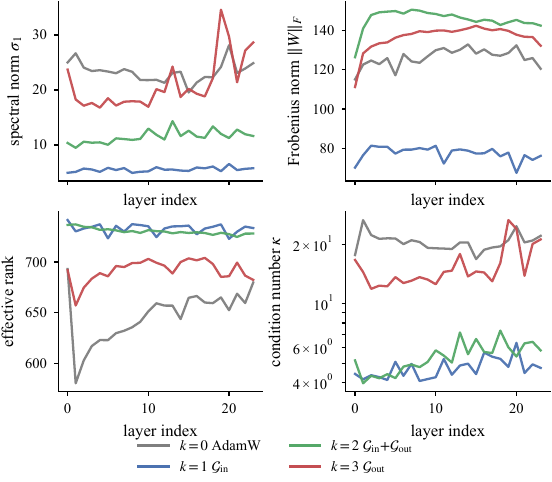}
  \caption{Per-layer spectral profile for $\mathcal{G}_{\mathrm{in}}$
    at $T_{\max}$, seed~$42$.}
  \label{fig:spectral-layer-gin}
\end{figure}

\begin{figure}[t]
  \centering
  \includegraphics[width=\linewidth]{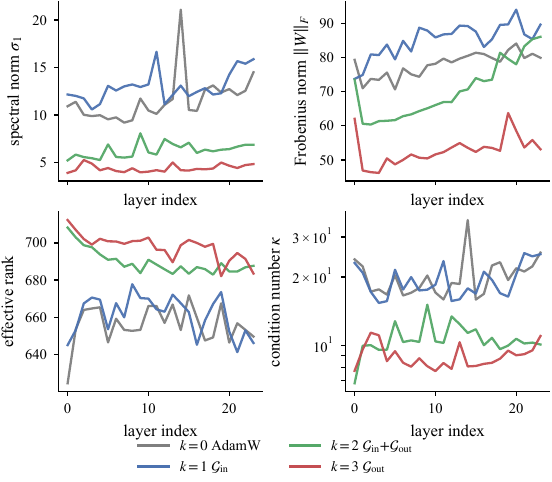}
  \caption{Per-layer spectral profile for $\mathcal{G}_{\mathrm{out}}$
    at $T_{\max}$, seed~$42$.}
  \label{fig:spectral-layer-gout}
\end{figure}

\paragraph*{Per-layer profiles.}
The per-layer profiles are consistent across all 24 layers with no
systematic depth dependence, and the half-budget profiles are
visually indistinguishable from the final ones; the spectral
structure thus stabilizes from mid-training onward
(Figures~\ref{fig:spectral-layer-gin},~\ref{fig:spectral-layer-gout}).
Under $k=1$, the $\mathcal{G}_{\mathrm{in}}$ spectral norm is
uniformly low (${\approx}5$--$6$) across layers, compared with
${\approx}22$--$28$ under AdamW.
Under $k=3$, the $\mathcal{G}_{\mathrm{out}}$ spectral norm is
uniformly low (${\approx}4$--$5$), compared with ${\approx}10$--$21$
under AdamW, with a notable spike at layer~14 in the AdamW and
$k=1$ baselines that is absent under $k=2$ and $k=3$.

\begin{figure}[t]
  \centering
  \includegraphics[width=\linewidth]{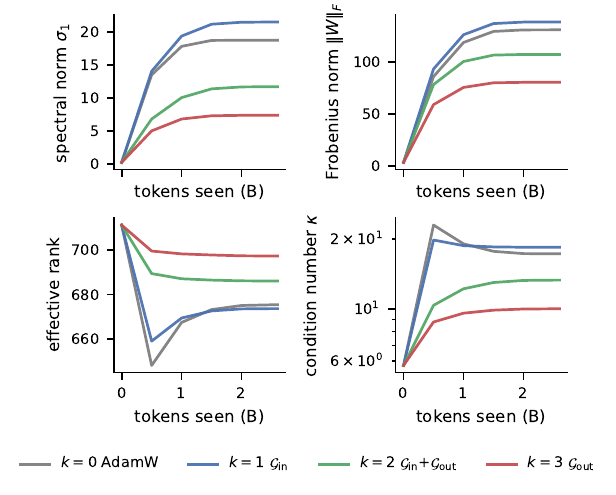}
  \caption{Layer-averaged $\mathcal{G}_{\mathrm{out}}$ spectral metrics
    over training, OpenWebText at $2.6\times 10^{9}$ tokens, seed~$42$.}
  \label{fig:sp-traj-owt26}
\end{figure}

\begin{figure}[t]
  \centering
  \includegraphics[width=\linewidth]{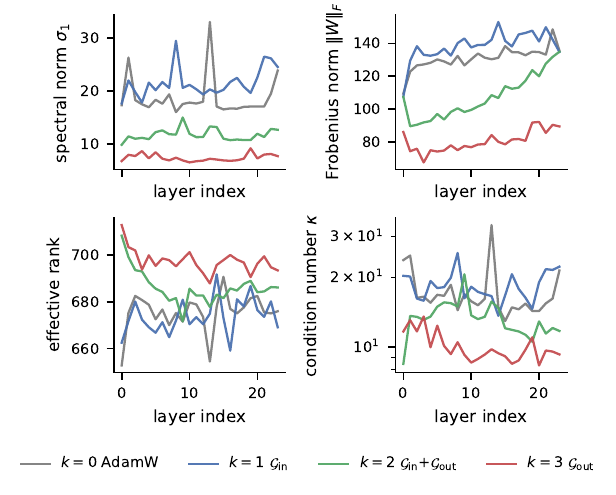}
  \caption{Per-layer $\mathcal{G}_{\mathrm{out}}$ spectral profile at the
    final checkpoint, OpenWebText at $2.6\times 10^{9}$ tokens,
    seed~$42$.}
  \label{fig:sp-layer-owt26}
\end{figure}

\begin{figure}[t]
  \centering
  \includegraphics[width=\linewidth]{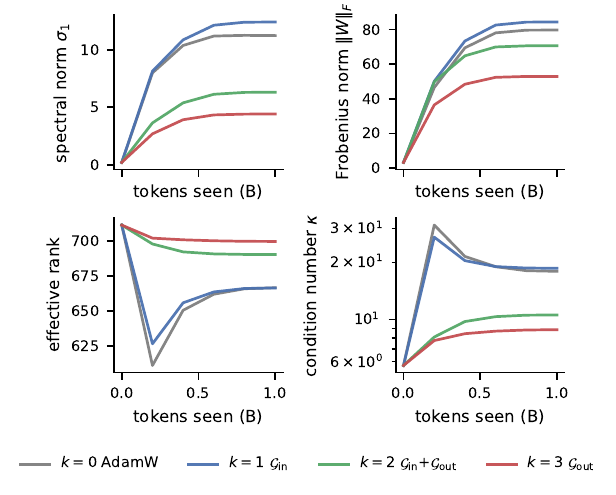}
  \caption{Layer-averaged $\mathcal{G}_{\mathrm{out}}$ spectral metrics
    over training, FineWeb-Edu at $10^{9}$ tokens, seed~$42$.}
  \label{fig:sp-traj-fw1b}
\end{figure}

\begin{figure}[t]
  \centering
  \includegraphics[width=\linewidth]{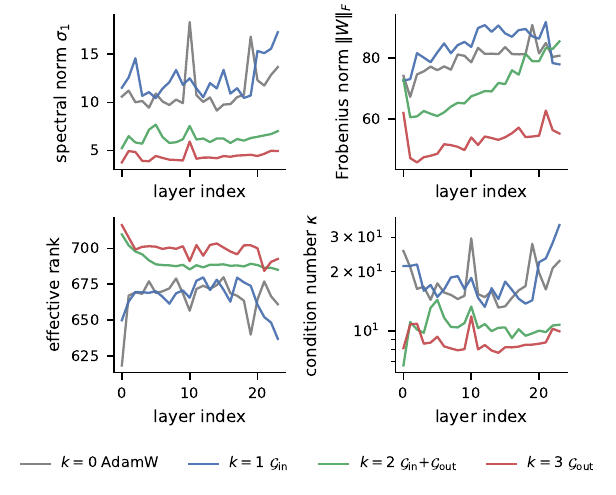}
  \caption{Per-layer $\mathcal{G}_{\mathrm{out}}$ spectral profile at the
    final checkpoint, FineWeb-Edu at $10^{9}$ tokens, seed~$42$.}
  \label{fig:sp-layer-fw1b}
\end{figure}

\begin{figure}[t]
  \centering
  \includegraphics[width=\linewidth]{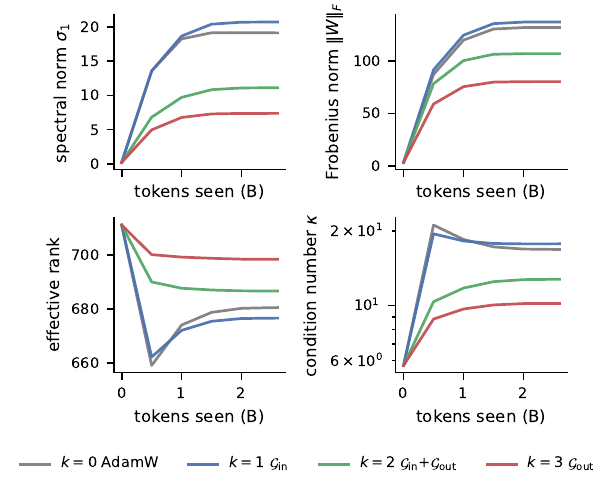}
  \caption{Layer-averaged $\mathcal{G}_{\mathrm{out}}$ spectral metrics
    over training, FineWeb-Edu at $2.6\times 10^{9}$ tokens, seed~$42$.}
  \label{fig:sp-traj-fw26}
\end{figure}

\begin{figure}[t]
  \centering
  \includegraphics[width=\linewidth]{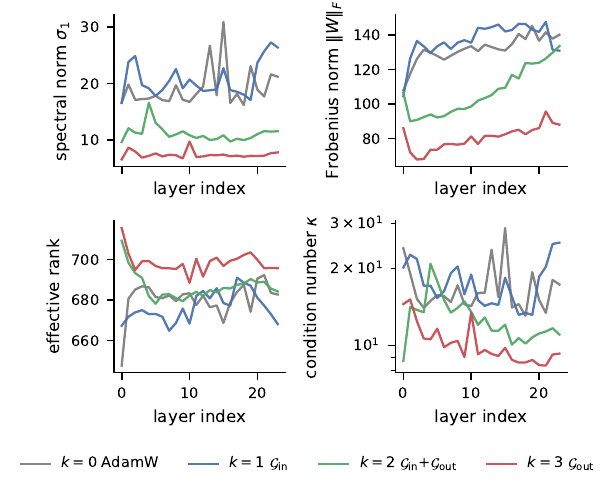}
  \caption{Per-layer $\mathcal{G}_{\mathrm{out}}$ spectral profile at the
    final checkpoint, FineWeb-Edu at $2.6\times 10^{9}$ tokens,
    seed~$42$.}
  \label{fig:sp-layer-fw26}
\end{figure}

\paragraph*{Robustness of the conditioning result.}
We recomputed the same $\mathcal{G}_{\mathrm{out}}$ diagnostics
(seed~$42$) for the three remaining corpus--budget cells of
Section~\ref{sec:results}, in the same format as the OpenWebText figures
of Section~\ref{sec:spectral}
(Figures~\ref{fig:sp-traj-owt26}--\ref{fig:sp-layer-fw26}); as
before, only the output projection is reported.
The conditioning result reproduces in every cell: $k=3$ reaches the
lowest condition number on $\mathcal{G}_{\mathrm{out}}$. At $10^{9}$
tokens FineWeb-Edu ($\kappa = 8.84$) essentially matches OpenWebText
($\kappa = 8.95$, Table~\ref{tab:spectral}), and at $2.6\times 10^{9}$
tokens $\kappa \approx 10$ on both corpora.
In these cells $k=1$ (Muon on $\mathcal{G}_{\mathrm{in}}$ only) leaves
$\mathcal{G}_{\mathrm{out}}$ slightly worse-conditioned than the AdamW
baseline, and $k=2$ falls in between, mirroring the loss ordering
$k=3 < k=2 < k=1 < k=0$. The $k=3$ conditioning advantage thus holds on
both corpora and at both budgets.

\begin{figure}[t]
  \centering
  \includegraphics[width=\linewidth]{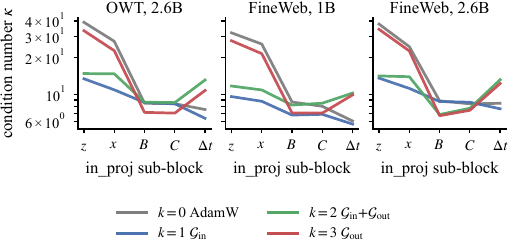}
  \caption{Layer-averaged condition number $\kappa$ of the five
    \texttt{in\_proj} row-slices ($z$, $x$, $B$, $C$, $\Delta t$) at the
    final checkpoint, by regime and corpus--budget cell (seed~$42$).
    Muon on \texttt{in\_proj} ($k=1$, $k=2$) conditions the ill-posed
    $z$ and $x$ slices but not the already-isotropic $B$, $C$,
    $\Delta t$.}
  \label{fig:subblock-slices}
\end{figure}

\paragraph*{Per-slice conditioning of \texttt{in\_proj}.}
Resolving \texttt{in\_proj} into its five row-slices makes the
heterogeneity behind Section~\ref{ssec:subblocks} explicit
(Figure~\ref{fig:subblock-slices}). Under AdamW ($k=0$) the slices span
an order of magnitude in conditioning: the gating channel $z$ and the
data channel $x$ are poorly conditioned ($\kappa\approx25$--$40$), while
$B$, $C$, and $\Delta t$ are near-isotropic ($\kappa\approx6$--$9$).
Placing \texttt{in\_proj} under Muon ($k=1$, $k=2$) sharply conditions
the ill-posed slices: $z$ and $x$ fall to $\kappa\approx9$--$15$, while
the already-good slices stay essentially unchanged. But this does not
lower the loss ($k=1,k=2$ trail $k=3$), and the slice Muon conditions
most aggressively, the gating channel $z$, is also the one whose
orthogonalization is most harmful in the sub-block ablation
(Section~\ref{ssec:subblocks}). For \texttt{in\_proj}, conditioning and
loss are therefore decoupled: Muon's benefit comes from the homogeneous
output projection $\mathcal{G}_{\mathrm{out}}$, not from conditioning the
heterogeneous input slices.

\section{Localizing the input-projection effect}
\label{ssec:localization}

The input projection \texttt{in\_proj} ($\mathcal{G}_{\mathrm{in}}$) is a
single $3352\times 768$ matrix that row-stacks five functionally distinct
sub-matrices $W_z, W_x, W_B, W_C, W_{\Delta t}$. To find which slice carries
the (small) $k=1$ effect, we replace the global Newton--Schulz pass on the
merged matrix with a \emph{split} variant in which each slice runs its own
Newton--Schulz iteration, then run a leave-one-out (LOO) ablation: starting
from the split all-Muon assignment, one slice at a time is moved back to
AdamW while the other four stay on Muon (OWT, $10^{9}$ tokens, single seed).

\paragraph*{Splitting helps.}
Per-slice orthogonalization beats one global pass: the split all-Muon
assignment reaches a final validation loss $0.024$ below the merged
whole-matrix $k=1$ run, so the merged Newton--Schulz under-conditions the
heterogeneous slices.

\paragraph*{The data channel carries the effect.}
Table~\ref{tab:loo} reports the increase in final validation loss when each
slice is removed from the split assignment. Removing the data channel $W_x$
is by far the most damaging ($+0.040$, larger than the gap between split and
the merged $k=1$ run itself), whereas removing the gating channel $W_z$ is
essentially free ($+0.001$); $W_B, W_C, W_{\Delta t}$ sit in between. The
$\mathcal{G}_{\mathrm{in}}$ benefit is thus localized almost entirely to the
$x$ slice.

\begin{table}[t]
  \centering
  \caption{Leave-one-out localization within \texttt{in\_proj}: increase in
    final validation loss when one slice is moved back to AdamW, relative to
    the split all-Muon assignment (OWT, $10^{9}$ tokens, single seed). The
    merged whole-matrix $k=1$ run is shown for reference.}
  \label{tab:loo}
  \footnotesize
  \begin{tabular}{l c}
    \hline
    Slice moved back to AdamW & $\Delta\mathcal{L}_{\mathrm{val}}$ \\
    \hline
    $-\,W_z$ (gate)           & $+0.001$ \\
    $-\,W_B$                  & $+0.008$ \\
    $-\,W_C$                  & $+0.008$ \\
    $-\,W_{\Delta t}$         & $+0.017$ \\
    $-\,W_x$ (data channel)   & $+0.040$ \\
    \hline
    merged $k{=}1$ (reference) & $+0.024$ \\
    \hline
  \end{tabular}
\end{table}

\begin{figure*}[t]
  \centering
  \includegraphics[width=\linewidth]{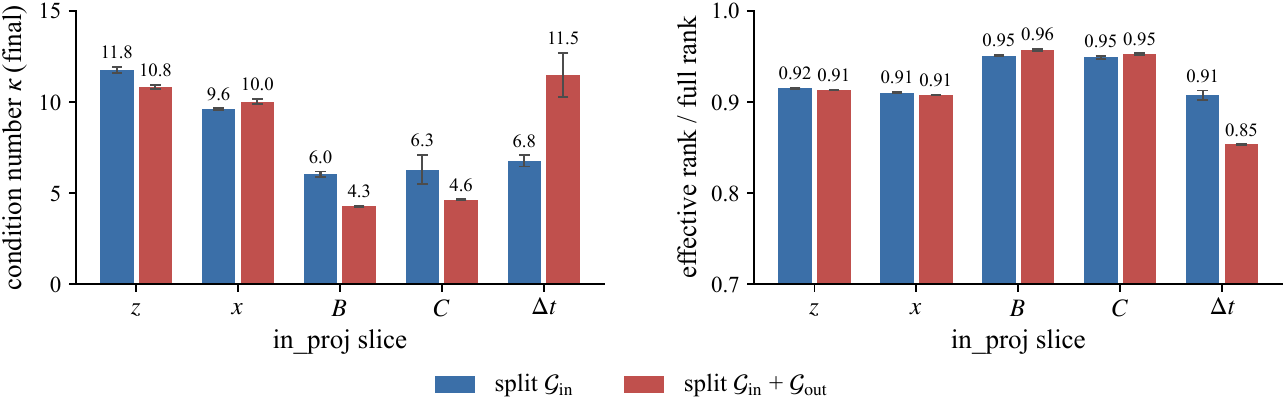}
  \caption{Per-slice condition number $\kappa$ (left) and effective-rank
    fraction (right) of the five \texttt{in\_proj} slices under split-Muon
    (OWT, $2.6\times 10^{9}$ tokens, mean over seeds). The gate $W_z$ and
    the data channel $W_x$ are the two worst-conditioned slices yet have
    opposite usefulness (removing $W_z$ is free, removing $W_x$ most
    costly), so conditioning does not predict which slice matters.}
  \label{fig:per-slice}
\end{figure*}

\paragraph*{Ill-conditioned $\neq$ useful.}
The per-slice spectra make this clear
(Figure~\ref{fig:per-slice}). Under split-Muon the gate $W_z$ and the data
channel $W_x$ are the two worst-conditioned slices, at a similar
$\kappa\approx 10$--$12$ and similar effective rank, yet their
contributions are opposite: removing $W_z$ is free, while removing $W_x$
is the most costly of the five. The better-conditioned slices $W_B$,
$W_C$, and $W_{\Delta t}$ contribute little. A slice's conditioning thus
does not predict its contribution, echoing the matrix-level finding of
Section~\ref{sec:spectral}: Muon's gain comes from the matrices (and
slices) the model actually \emph{uses}, not from those it conditions
most.

\paragraph*{The output projection remains the driver.}
Localizing the $\mathcal{G}_{\mathrm{in}}$ effect does not make it
competitive with $\mathcal{G}_{\mathrm{out}}$. Stacking the best
$\mathcal{G}_{\mathrm{in}}$ recipe on top of the output projection (split
$\mathcal{G}_{\mathrm{in}}$ \emph{and} Muon on $\mathcal{G}_{\mathrm{out}}$)
does not beat $k=3$ alone: on OWT at $2.6\times 10^{9}$ tokens the two tie
on validation, with $\mathcal{G}_{\mathrm{out}}$-only marginally ahead
($3.118$ vs.\ $3.128$; hyperparameter-matched runs, validation only). The
split analysis tells us \emph{where} the $\mathcal{G}_{\mathrm{in}}$ effect
lives, but it is a mechanistic refinement, not a better model:
$\mathcal{G}_{\mathrm{out}}$ stays the single best target.

\bibliographystyle{IEEEtran}
\bibliography{references}

\end{document}